\documentclass[10pt]{article}
\usepackage{geometry} 
\usepackage{natbib, algorithm, algorithmic,hyperref}

\usepackage{microtype}
\usepackage{graphicx}
\usepackage{subcaption}
\usepackage{booktabs} 

\usepackage{graphicx} 
\usepackage{grffile}

\usepackage{hyperref}

\usepackage{amsmath,amsfonts,bm,mathtools}

\newcommand{\eqdef}{\coloneqq}

\newcommand{\cO}{{\cal O}}
\newcommand{\cC}{{\cal C}}
\newcommand{\cN}{{\cal N}}
\newcommand{\cD}{{\cal D}}

\newcommand{\R}{\mathbb{R}}

\newcommand{\E}{\mathbb{E}}
\newcommand{\B}{\mathbb{B}}
\newcommand{\U}{\mathbb{U}}
\newcommand{\sphere}{\mathbb{S}}

\def\Pr{{\rm Prob}}

\def\supp{{\rm supp}}

\DeclareMathOperator*{\sign}{{\rm sign}}

\DeclareMathOperator*{\dist}{{\rm dist}}

\DeclarePairedDelimiter\ceil{\lceil}{\rceil}

\def\<{\left\langle}
\def\>{\right\rangle}
\def\[{\left[}
\def\]{\right]}
\def\({\left(}
\def\){\right)}

\usepackage[dvipsnames]{xcolor}
\usepackage{pifont}

\usepackage{nicefrac}
\usepackage{xcolor}
\graphicspath{{plots/}}

\newtheorem{theorem}{Theorem}

\newtheorem{remark}{Remark}

\newtheorem{lemma}{Lemma}
\newtheorem{definition}{Definition}

\begin{document}

\title{Uncertainty Principle for Communication Compression  in Distributed and Federated Learning and the Search for an Optimal Compressor}

\author{Mher Safaryan \qquad Egor Shulgin\footnote{Part of this work was done by the author as a student at the Moscow Institute of Physics and Technology, Dolgoprudny, Russia.} \qquad Peter Richt\'arik\\
\phantom{XXX} \\
\em King Abdullah University of Science and Technology\\
\em Thuwal, Saudi Arabia
}

\date{26 / 1 / 2021}

\maketitle

\begin{abstract}
In order to mitigate the high communication cost in  distributed and federated learning, various vector compression schemes, such as quantization, sparsification and dithering, have become very popular. In designing a compression method, one aims to communicate as few bits as possible, which minimizes the cost per communication round, while at the same time attempting to impart as little distortion (variance) to the communicated messages as possible, which minimizes the adverse effect of the compression on the overall number of communication rounds. However, intuitively, these two goals are fundamentally in conflict: the more compression we allow, the more distorted the messages become. We formalize this intuition and prove an {\em uncertainty principle} for randomized compression operators, thus quantifying this limitation mathematically, and {\em effectively providing asymptotically tight lower bounds on what might be achievable with communication compression}. Motivated by these developments, we call for the search for the optimal compression operator. In an attempt to take a first step in this direction, we consider an unbiased compression method inspired by the Kashin representation of vectors, which we call {\em Kashin compression (KC)}. In contrast to all previously proposed compression mechanisms, KC enjoys a {\em dimension independent} variance bound for which we derive an explicit formula even in the regime when only a few bits need to be communicate per each vector entry.
\end{abstract}


 \tableofcontents
 
 
\section{Introduction}

In the quest for high accuracy machine learning models, both the size of the model and consequently the amount of data necessary to train the model have been hugely increased over time \citep{Sch,Vaswani2019-overparam}.
Because of this, performing the learning process on a single machine is often  infeasible. In a typical scenario of distributed learning, the training data (and possibly the model as well) is spread across different machines and thus the process of training is done in a distributed manner \citep{bekkerman2011scaling,vogels}.
Another scenario, most common to federated learning \citep{FEDLEARN,FL2017-AISTATS,Karimireddy2019}, is when training data is inherently distributed across a large number of mobile edge devices due to data privacy concerns.

\subsection{Communication bottleneck}
In all cases of distributed learning and federated learning, information (e.g. current stochastic gradient vector or current state of the model) communication between computing nodes is inevitable, which forms the primary bottleneck of such systems \citep{ZLKALZ, LHMWD}. This issue is especially apparent in federated learning, where computing nodes are devices with essentially inferior power and the network bandwidth is considerably slow \citep{FL_survey_2019}.

There are two general approaches to address/tackle this problem. One line of research dedicated to so-called local methods suggests to do more computational work before each communication in the hope that those would increase the worth/impact/value of the information to be communicated \citep{Goyal2017:large,tonko,Stich-localSGD,localSGD-AISTATS2020}. An alternative approach investigates inexact/lossy information compression strategies which aim to  send  approximate but relevant information encoded with less number of bits.  In this work we focus on the second approach of {\em compressed learning}. Research in this latter stream splits into two orthogonal directions. To explore savings in communication, various (mostly randomized) compression operators have been proposed and analyzed such as random sparsification \citep{RDME,tonko}, Top-$k$ sparsification \citep{alistarh2018convergence}, standard random dithering \citep{Goodall1951:randdithering,Roberts1962:randdithering,AGLTV}, natural dithering \citep{Cnat}, ternary quantization \citep{terngrad}, and sign quantization \citep{SQSM,BWAA,BZAA,LCCH,sign_descent_2019}. Table~\ref{table:comp-methods} summarizes the most common compression methods with their variances and the number of encoding bits.

In designing a compression operator, one aims to (i) encode the compressed information with as few bits as possible, which minimizes the cost per communication round, and (ii) introduce as little noise (variance) to the communicated messages as possible, which minimizes the adverse effect of the compression on the overall iteration complexity.

\begin{table*}
\caption{Compression operators in $\U(\omega)$ and $\B(\alpha)$ with dimension $d$. The number of encoding bits for KC depends on the quantization operator. Mentioned formula for KC uses ternary quantization.}
\label{table:comp-methods}
\begin{center}
\begin{small}
\begin{sc}
\begin{tabular}{l@{\hskip 0.03in}c@{\hskip 0.03in}r@{\hskip 0.03in}l@{\hskip 0.03in}c@{\hskip 0.1in}c}
\toprule
Compression Method & Unbiased? & Variance & $\omega$ & Variance $\alpha$ & Bits $b$ (in \textit{binary32}) \\
\midrule
Random sparsification   & {\color{PineGreen} yes} & $\tfrac{d}{k}-1$ & $\approx \cO(\tfrac{d}{k})$ & & $32k + \log_2\binom{d}{k}$ \\
Top-$k$ sparsification  & {\color{Red} no} & & & $1 -\tfrac{k}{d}$ & $32k + \log_2\binom{d}{k}$ \\
Standard Dithering      & {\color{PineGreen} yes} & $\min(\tfrac{\sqrt{d}}{s}, \tfrac{d}{s^2})$ & $\approx \cO(\tfrac{\sqrt{d}}{s})$ & & $\cO\(s(s+\sqrt{d})\)$ \\
Natural Dithering       & {\color{PineGreen} yes} & $\min(\tfrac{\sqrt{d}}{2^{s-1}}, \tfrac{d}{2^{2-2s}})$ & $\approx \cO(\tfrac{\sqrt{d}}{2^{s-1}})$ & & $31 + d \log_2(2s+1)$ \\
Ternary Quantization    & {\color{PineGreen} yes} & $\sqrt{d}-1$ & $\approx \cO(\sqrt{d})$ & & $31 + d\log_2 3$ \\
Scaled Sign Quantization& {\color{Red} no} & & & $1-\tfrac{1}{d}$ & $31 + d$\\
\textbf{Kashin Compression (new)} & {\color{PineGreen} yes} & $\(\frac{10\sqrt{\lambda}}{\sqrt{\lambda}-1}\)^4$ & $\approx \cO(1)$ & & $31+\log_2 3\cdot\lambda d$ \\
\bottomrule
\end{tabular}
\end{sc}
\end{small}
\end{center}
\vskip -0.1in
\end{table*}

\subsection{Compressed learning}

In order to utilize these compression methods efficiently, a lot of research has been devoted to the study of learning algorithms with compressed communication. Obviously, the presence of compression in a learning algorithm affects the training process and since compression operator encodes the original information approximately, it should be anticipated to increase the number of communication rounds. Table~\ref{table:iter-complexity} highlights four  gradient-type compressed learning algorithms with their corresponding setup and iteration complexity: \begin{itemize}
\item [(i)] distributed Gradient Descent (GD) with compressed gradients \citep{KFJ}, 
\item [(ii)] distributed Stochastic Gradient Descent (SGD) with gradient quantization and compression variance reduction \citep{DIANA-VR},  
\item [(iii)] distributed SGD with bi-directional gradient compression \citep{Cnat}, and 
\item [(iv)] distributed SGD with gradient compression and twofold error compensation \citep{DoubleSqueeze2019}. 
\end{itemize}

\begin{table*}
\caption{Iteration complexities of different learning algorithms with respect to the variance ($\omega$ or $\alpha$) of compression operator. For strongly convex problems $\kappa$ is the condition number, $\epsilon$ is the accuracy and $n$ is the number of nodes in distributed setup.}
\label{table:iter-complexity}
\begin{center}
\begin{small}
\begin{sc}
\begin{tabular}{lcc}
\toprule
Optimization Algorithm & Objective Function & Iteration complexity  \\
\midrule
Compressed GD \citep{KFJ} & smooth, strongly convex   & $ \cO\( \kappa (\omega+1) \log\frac{1}{\varepsilon} \)$ \\
DIANA \citep{DIANA-VR}  & smooth, strongly convex         & $\cO\( (\kappa + \omega\frac{\kappa}{n} + \omega) \log\frac{1}{\varepsilon} \)$ \\
Distributed SGD \citep{Cnat} & smooth, non-convex   & $\cO\( (\omega+1)^2 \frac{1}{\varepsilon^2} \)$ \\
DoublSqueeze \citep{DoubleSqueeze2019}                 & smooth, non-convex      & $\cO\( \frac{1}{n\varepsilon^2} + \frac{1}{1-\alpha}\frac{1}{\varepsilon^{1.5}} + \frac{1}{\varepsilon} \)$ \\
\bottomrule
\end{tabular}
\end{sc}
\end{small}
\end{center}
\vskip -0.1in
\end{table*}

In all cases, the iteration complexity depends on the variance of the underlying compression scheme and grows as more compression is applied. For this reason, we are interested in compression methods which save in communication by using less bits and minimize iteration complexity by introducing lower variance. However, intuitively and also evidently from Table~\ref{table:comp-methods}, these two goals are in fundamental conflict, i.e. requiring fewer bits to be communicated in each round introduces higher variance, and demanding small variance forces  more bits to be communicated.

\begin{figure}[ht]
\begin{center}
\centerline{\includegraphics[width=0.7\linewidth]{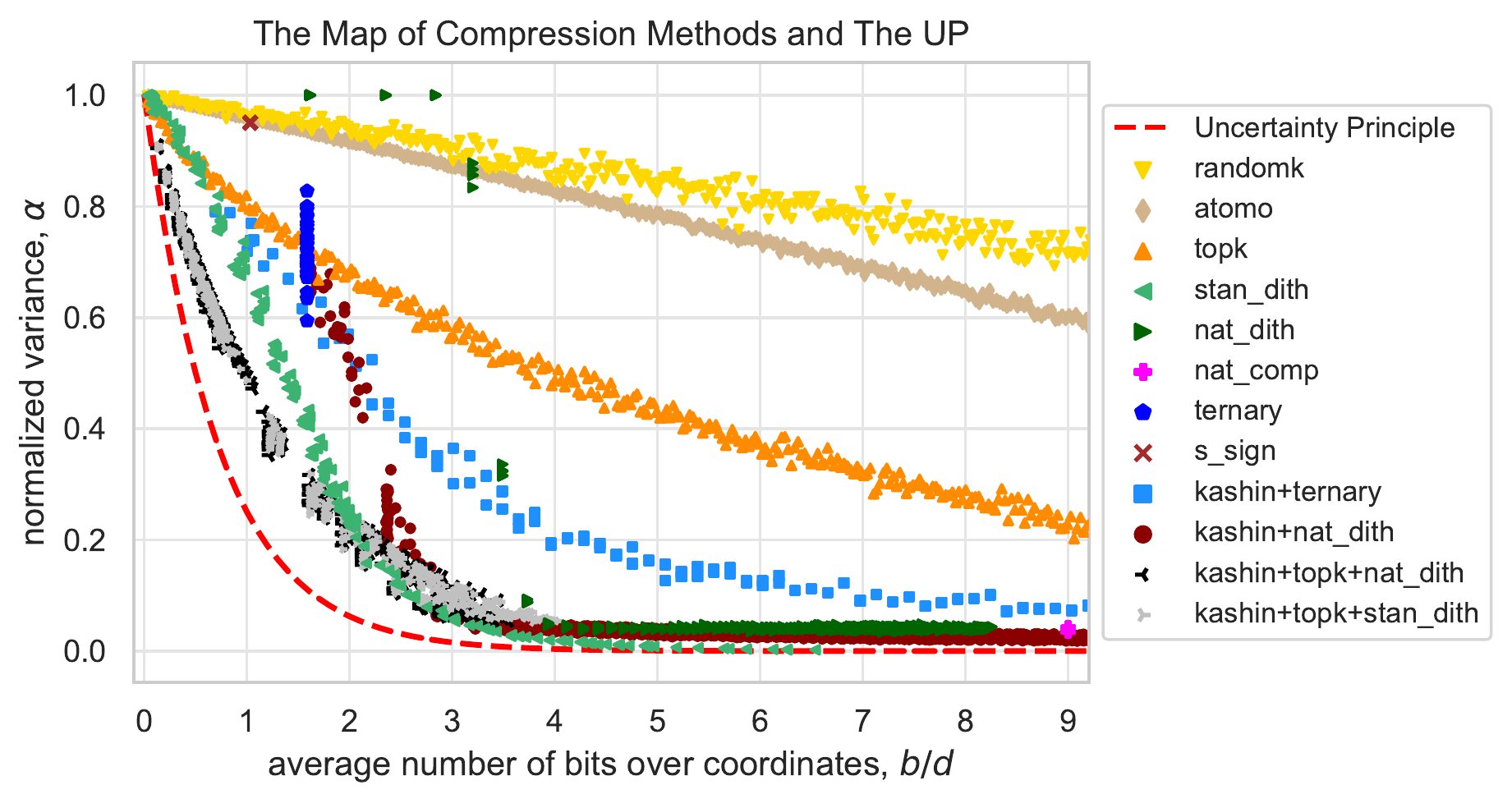}}
\caption{Comparison of the most common compression methods based on their normalized variance $\alpha\in[0,1]$ and the average number of encoding bits per coordinate. Each color represents one compression method, each marker indicates one particular $d=10^3$ dimensional vector randomly generated from Gaussian distribution, which subsequently gets compressed by the compression operator mentioned in the legend. Dashed red line shows the lower of bound of the uncertainty principle (\ref{ups}). The motivation for using Gaussian random vectors stems from the fact that it is the hardest source to encode.
}
\label{fig:var_bits}
\end{center}
\vskip -0.1in
\end{figure}

\subsection{Contributions}

We make the following contributions in this work:

{\bf Uncertainty Principle.} We formalize this intuitive trade-off and {\em prove an uncertainty principle for randomized compression operators}, which quantifies this limitation mathematically with the inequality
\begin{equation}\label{ups}
\alpha\cdot 4^{\nicefrac{b}{d}} \ge 1,
\end{equation}
where $\alpha\in[0,1]$ is the normalized variance (or contraction factor) associated with the compression operator (Definition~\ref{def:biased}),  $b$ is the number of bits required to encode the compressed vector and $d$ is the dimension of the vector to be compressed.
It is a universal property of compressed communication, completely independent of the optimization algorithm and the problem that distributed training is trying to solve.
We visualize this principle in Figure~\ref{fig:var_bits}, where many possible combinations of parameters $\alpha$ and $\nicefrac{b}{d}$ were computed for various compression methods. The dashed red line, indicating the lower bound (\ref{ups}), bounds all possible combinations of all compression operators, thus validating the obtained uncertainty principle for randomized compression operators. We also show that this lower bound is asymptotically tight.

{\bf Kashin Compression.} Motivated by this principle, we then focus on the search for the optimal compression operator. In an attempt to take a first step in this direction, we investigate an unbiased compression operator inspired by  Kashin representation of vectors~\citep{kashin}, which we call Kashin Compression (KC).
In contrast to all previously proposed compression methods, KC enjoys a {\em dimension independent variance bound } even in a severe compression regime when only a few bits per coordinate can be communicated. We derive an explicit formula for the variance bound.
Furthermore, we experimentally observed the superiority of KC in terms of communication savings and stabilization property when compared against commonly used compressors proposed in the literature. In particular, Figure~\ref{fig:var_bits} justifies that KC combined with Top-$k$ sparsification and dithering operators yields a compression method which is the most efficient scheme when communication is scarce.
Kashin's representation has been used heuristically in the context of federated learning \citep{CKMT-fedkashin} to mitigate the communication cost.
We believe KC should be of high interest in federated and distributed learning.

\section{Uncertainty principle for compression operators}\label{sec:up}

In general, an uncertainty principle refers to any type of mathematical inequality expressing some fundamental trade-off between two measurements. The classical Heisenberg's uncertainty principle in quantum mechanics~\citep{Heisenberg1927} shows the trade-off between the position and momentum of a particle. In harmonic analysis, the uncertainty principle limits the localization of values of a function and its Fourier transform at the same time~\citep{UP-Harmonic1994}. Alternatively in the context of signal processing, signals cannot be simultaneously localized in both time domain and frequency domain~\citep{Gabor1946}. The uncertainty principle in communication deals with the quite intuitive trade-off between information compression (encoding bits) and approximation error (variance), namely more compression forces heavier distortion to communicated messages and tighter approximation requires less information compression.

In this section, we present our UP for communication compression revealing the trade-off between encoding bits of compressed information and the variance produced by compression operator. First, we describe UP for some general class of biased compressions. Afterwards, we specialize it to the class of unbiased compressions.

\subsection{UP for biased compressions}
We work with the class of biased compression operators which are contractive.

\begin{definition}[Biased Compressions]\label{def:biased}
Let $\B(\alpha)$ be the class of biased (and possibly randomized) compression operators $\mathcal{C}\colon\R^d\to\R^d$ with $\alpha\in[0,1]$ contractive property, i.e. for any $x\in\R^d$
\begin{equation}\label{class-biased}
\E\[\|\mathcal{C}(x)-x\|_2^2\] \le \alpha\|x\|_2^2.
\end{equation}
\end{definition}

The parameter $\alpha$ can be seen as the normalized variance of the compression operator. Note that the compression $\mathcal{C}$ does not need to be randomized to belong to this class. For instance, Top-$k$ sparsification operator satisfies (\ref{class-biased}) without the expectation for $\alpha = 1-\frac{k}{d}$. Next, we formalize uncertainty principle for the class $\B(\alpha)$.

\begin{theorem}\label{thm:biased_up}
Let $\mathcal{C}\colon \R^d\to\R^d$ be any compression operator from $\B(\alpha)$ and $b$ be the total number of bits needed to encode the compressed vector $\mathcal{C}(x)$ for any $x\in\R^d$. Then the following form of uncertainty principle holds
\begin{equation}\label{biased_up}
\alpha \cdot 4^{\nicefrac{b}{d}} \ge 1.
\end{equation}
\end{theorem}
One can view the {\em binary32} and {\em binary64} floating-points formats as biased compression methods for the actual real numbers (i.e. $d=1$), using only 32 and 64 bits respectively to represent a single number. Intuitively, these formats have their precision (i.e. $\sqrt{\alpha}$) limits and the uncertainty principle (\ref{biased_up}) shows that the precision cannot be better than $2^{-32}$ for {\em binary32} format and $2^{-64}$ for {\em binary64} format. Thus, any floating-point format representing a single number with $r$ bits has precision constraint of $2^{-r}$, where the base $2$ stems from the binary nature of the bit.

Furthermore, notice that compression operators can achieve zero variance in some settings, e.g. ternary or scaled sign quantization when $d=1$ (see Table \ref{table:comp-methods}). On the other hand, the UP (\ref{biased_up}) implies that the normalized variance $\alpha>0$ for any finite bits $b$. The reason for this inconsistency comes from the fact that, for instance, the {\em binary32} format encodes any number with 32 bits and the error $2^{-32}$ is usually ignored in practice.  We can adjust UP to any digital format, using $r$ bits per single number, as
\begin{equation}\label{biased_up_adjusted}
\(\alpha + 4^{-r}\) 4^{\nicefrac{b}{d}} \ge 1.
\end{equation}

\subsection{UP for unbiased compressions}

We now specialize (\ref{biased_up}) to the class of unbiased compressions. First, we recall the definition of unbiased compression operators with a given variance.

\begin{definition}[Unbiased Compressions]
Denote by $\U(\omega)$ the class of unbiased compression operators $\mathcal{C}\colon\R^d\to\R^d$ with variance $\omega>0$, that is, for any $x\in\R^d$
\begin{equation}\label{class-unbiased}
\textstyle \E\[\mathcal{C}(x)\] = x, \qquad \E\[\|\mathcal{C}(x)-x\|_2^2\] \le \omega\|x\|_2^2.
\end{equation}
\end{definition}

To establish an uncertainty principle for $\cC\in \U(\omega)$, we show that all unbiased compression operators with the proper scaling factor are included in $\B(\alpha)$.
\begin{lemma}\label{lem:inclusion}
If $\mathcal{C}\in\U(\omega)$, then $\tfrac{1}{\omega+1}\mathcal{C}\in\B(\tfrac{\omega}{\omega+1})$.
\end{lemma}

Using this inclusion, we can apply Theorem \ref{thm:biased_up} to the class $\U(\omega)$ and derive an uncertainty principle for unbiased compression operators.

\begin{theorem}\label{thm:unbiased_up}
Let $\mathcal{C}\colon \R^d\to\R^d$ be any unbiased compression operator with variance $\omega\ge 0$ and $b$ be the total number of bits needed to encode the compressed vector $\mathcal{C}(x)$ for any $x\in\R^d$. Then the uncertainty principle takes the  form
\begin{equation}\label{unbiased_up}
\frac{\omega}{\omega+1} \cdot 4^{\nicefrac{b}{d}} \ge 1.
\end{equation}
\end{theorem}

\section{Compression with polytopes}\label{sec:polytop}

Here we describe an unbiased compression scheme based on polytopes. With this particular compression we illustrate that it is possible for unbiased compressions to have dimension independent variance bounds and at the same time communicate a single bits per coordinate.

Let $x\in\R^d$ be nonzero vector that we need to encode. First, we project the vector on the unit sphere $$\sphere^{d-1} = \{x\in\R^d\colon \|x\|_2=1\},$$ thus separating the magnitude $\|x\|_2\in\R$ from the direction $x/\|x\|_2\in\sphere^{d-1}$. The magnitude is a dimension independent scalar value and we can transfer it cheaply, say by 32 bits. To encode the unit vector $x/\|x\|_2$ we approximate the unit sphere by polytopes and then randomize over the vertices of the  polytope. Polytopes can be seen as generalizations of planar polygons in high dimensions. Formally, let $P_m$ be a polytope with vertices $\{v_1,v_2,\dots,v_m\}\subset\R^d$ such that it contains the unit sphere, i.e. $\sphere^{d-1}\subset P_m$, and all vertices are on the sphere of radius $R>1$. Then, any unit vector $v\in\sphere^{d-1}$ can be expressed as a convex combination $\sum_{k=1}^m w_k v_k$ with some non-negative weights $w_k=w_k(x)$. Equivalently, $v$ can be expressed as an expectation of a random vector over $v_k$ with probabilities $w_k$. Therefore, the direction $x/\|x\|_2$ could be encoded with roughly $\log m$ bits and the variance $\omega$ of compression will depend on the approximation, more specifically $\omega = R^2 - 1$. In \cite{Koch1, Koch2} it is given a constructive proof on approximation of the $d$-dimensional unit sphere by polytopes with $m\ge 2d$ vertices for which $\omega = \cO\(\tfrac{d}{\log\nicefrac{m}{d}}\)$. So, choosing the number of vertices to be $m = 2^d$, one gets an unbiased compression operator with $\cO(1)$ variance (independent of dimension $d$) and with 1 bit per coordinate encoding.

This simple method does not seem to be practical as $2^d$ vertices of the polytope either need to be stored or computed  each time they are used, which is infeasible for large dimensions. However, we use this construction and show that lower bounds (\ref{unbiased_up}) and hence (\ref{biased_up}) are asymptotically tight.
\begin{theorem}\label{asym-tightness}
For any $\epsilon>0$ there exists an unbiased compression $\cC\colon\sphere^{d-1}\to\R^d$ of some dimension $d$ and variance $\omega>0$ such that
\begin{equation}\label{epsilon-optimal-bound}
\frac{\omega}{\omega+1} 4^{\nicefrac{b}{d}} < 1 + \epsilon,
\end{equation}
where $b$ is the number of bits to encode $\cC(x)\in\R^d$ for any $x\in\sphere^{d-1}$.
\end{theorem}

\section{Compression with Kashin's representation}\label{sec:kashin_repr}
In this section we review the notion of Kashin's representation, the algorithm in \cite{LyVe} on computing it efficiently and then describe the quantization step.
\subsection{Representation systems}
The most common way of compressing a given vector $x\in\R^d$ is to use its {\em orthogonal representation} with respect to the standard basis $(e_i)_{i=1}^d$ in $\R^d$:
\begin{equation*}
 x = \sum \limits_{i=1}^d x_i e_i, \qquad x_i = \<x, e_i\>.
\end{equation*}
However, the restriction of orthogonal expansions is that coefficients $x_i$ are independent in the sense that if we lost one of them, then we cannot recover it even approximately. Furthermore, each coefficient $x_i$ may carry very different portion of the total information that vector $x$ contains; some coefficients may carry more information than others and thus be more sensitive to compression.

For this reason, it is preferable to use {\em tight frames} and {\em frame representations} instead. Tight frames are generalizations of orthonormal bases, where the system of vectors are not required to be linearly independent. Formally, vectors $(u_i)_{i=1}^D$ in $\R^d$ form a tight frame if any vector $x\in\R^d$ admits a frame representation
\begin{equation}\label{frame_repr}
 x = \sum \limits_{i=1}^D a_i u_i, \qquad a_i = \<x, u_i\>.
\end{equation}
Clearly, if $D>d$ (the case we are interested in), then the system $(u_i)_{i=1}^D$ is linearly dependent and hence the representation (\ref{frame_repr}) with coefficients $a_i$ is not unique. The idea is to exploit this redundancy and choose coefficients $a_i$ in such a way to spread the information uniformly among these coefficients. However, the frame representation may not distribute the information well enough. Thus, we need a particular representation for which coefficients $a_i$ have smallest possible dynamic range.

For a frame $(u_i)_{i=1}^D$ define the $d\times D$ {\em frame matrix} $U$ by stacking frame vectors $u_i$ as columns. It can be easily seen that being a tight frame is equivalent to frame matrix to be orthogonal, i.e. $U U^\top = I_d$, where $I_d$ is the $d\times d$ identity matrix. Using the frame matrix $U$, frame representation (\ref{frame_repr}) takes the form $x = U a$.

\begin{definition}[Kashin's representation]
Let $(u_i)_{i=1}^D$ be a tight frame in $\R^d$. Define Kashin's representation of $x\in\R^d$ with level $K$ the following expansion
\begin{equation}\label{kashin_repr}
 x = \sum \limits_{i=1}^D a_i u_i, \qquad \max \limits_{1\le i\le D}|a_i| \le \frac{K}{\sqrt{D}} \|x\|_2.
\end{equation}
\end{definition}

\textbf{Optimality.}
As noted in \citep{LyVe}, Kashin's representation has the smallest possible dynamic range $\nicefrac{K}{\sqrt{D}}$, which is $\sqrt{d}$ times smaller then dynamic range of the frame representation (\ref{frame_repr}).

\textbf{Existence.}
It turns out that not every tight frame can guarantee Kashin's representation with constant level. The following existence result is based on Kashin's theorem \citep{kashin}:
\begin{theorem}[\cite{kashin}]
There exist tight frames in $\R^d$ with arbitrarily small redundancy $\lambda = \nicefrac{D}{d} > 1$, and such that every vector $x\in\R^d$ admits Kashin's representation with level $K = K(\lambda)$ that depends on $\lambda$ only (not on $d$ or $D$).
\end{theorem}

\subsection{Computing Kashin's representation}
To compute Kashin's representation we use the algorithm developed in \cite{LyVe}, which transforms the frame representation (\ref{frame_repr}) into Kashin's representation (\ref{kashin_repr}). The algorithm requires tight frame with frame matrix satisfying the restricted isometry property:

\begin{algorithm}[tb]
   \caption{Computing Kashin's representation \citep{LyVe}}
   \label{alg:example}
\begin{algorithmic}\label{frame_to_kashin}
   \STATE {\bfseries Input:} orthogonal $d\times D$ matrix $U$ which satisfies RIP with parameters $\delta,\eta\in(0,1)$, a vector $x\in\R^d$ and a number of iterations $r$.
   \STATE {\bf Initialize} $a = 0 \in \R^D,\, M = \|x\|_2/\sqrt{\delta D}$.
   \STATE {\bfseries repeat} $r$ times
   \STATE $b = U^\top x$
   \STATE$\hat{b} = \sign(b) \cdot \min(|b|, M)$
   \STATE $x = x - U \hat{b}$
   \STATE $a = a + \hat{b}$
   \STATE$M = \eta M$
   \STATE {\bfseries return} $a$
   \STATE {\bfseries Output:} Kashin's coefficients of $x$ with level $K = 1/(\sqrt{\delta}(1-\eta))$ and with accuracy $\eta^r\|x\|_2$, i.e.
   $$ \left\|x - U a\right\|_2 \le \eta^r\|x\|_2,\qquad \max \limits_{1\le i\le D}|a_i| \le \frac{K}{\sqrt{D}} \|x\|_2.$$
\end{algorithmic}
\end{algorithm}

\begin{definition}[Restricted Isometry Property (RIP)]\label{def:up_matrix}
A given $d\times D$ matrix $U$ satisfies the Restricted Isometry Property with parameters $\delta,\eta\in(0,1)$ if for any $x\in\R^d$
\begin{equation}\label{up_matrix}
\textstyle |\supp(x)| \le \delta D \quad \Rightarrow \quad \|U x\|_2 \le \eta\|x\|_2.
\end{equation}
\end{definition}

In general, for an orthogonal $d\times D$ matrix $U$ we can only guarantee the inequality $\|U x\|_2 \le \|x\|_2$ if $x\in\R^d$. The RIP requires $U$ to be a contraction mapping for sparse $x$. With a frame matrix satisfying RIP, the analysis of Algorithm \ref{frame_to_kashin} from \citep{LyVe} yields a formula for the level of Kashin's representation:
\begin{theorem}[see Theorem 3.5 of \cite{LyVe}]
Let $(u_i)_{i=1}^D$ be a tight frame in $\R^d$ which satisfies RIP with parameters $\delta, \eta$. Then any vector $x\in\R^d$ admits a Kashin's representation with level
\begin{equation}\label{kashin_level}
 K = \frac{1}{\sqrt{\delta}(1-\eta)}.
\end{equation}
\end{theorem}

\subsection{Quantizing Kashin's representation}\label{subsec:quant-kashin}
We utilize Kashin's representation to design a compression method, which will enjoy dimension-free variance bound on the approximation error. Let $x\in\R^d$ be the vector that we want to communicate and $\lambda>1$ be the redundancy factor so that $D=\lambda d$ is positive integer. First we find Kashin's representation of $x$, i.e. $x = U a$ for some $a\in\R^D$, and then quantize coefficients $a_i$ using any unbiased compression operator $\cC\colon\R^D\to\R^D$ that preserves the sign and maximum magnitude:
\begin{equation}\label{quantizer_di}
\textstyle 0 \le \cC(a)\sign(a) \le \|a\|_{\infty}, \quad a\in\R^D.
\end{equation}
For example, ternary quantization or any dithering (standard random, natural) can be applied. The vector that we communicate is the quantized coefficients $\mathcal{C}(a)\in\R^D$ and KC is defined via
$$ \cC_{\kappa}(x) = U\cC(a).$$
Due to unbiasedness of $\cC$ and linearity of  expectation, we preserve unbiasedness for $\cC_{\kappa}$:
\begin{equation*}
\textstyle \E[\cC_{\kappa}(x)] = \E\left[U \cC(a) \right] = U \E\left[\cC(a) \right] = U a = x.
\end{equation*}

Then the error of approximation can be bounded uniformly (without the expectation) as follows
\begin{eqnarray*}
\|\cC_{\kappa}(x) - x\|_2^2 = \|U\cC(a) - U a\|_2^2 \le \|\cC(a) - a\|_2^2
&\le & D \max_{1\le i\le D}(\cC(a)_i - a_i)^2  \\
&\le & D \|a\|_{\infty}^2 \le  D \left(\frac{K(\lambda)}{\sqrt{D}} \|x\|_2 \right)^2 =  K^2(\lambda)\|x\|_2^2.
\end{eqnarray*}

The obtained uniform upper bound $K(\lambda)^2$ does not depend on the dimension $d$. It depends only on the redundancy factor $\lambda>1$ which should be chosen depending on how less we want to communicate. Thus, KC $\cC_{\kappa}$ with any unbiased quantization (\ref{quantizer_di}) belongs to $\U\(K^2(\lambda)\)$.

Described Kashin representation and Kashin compression has similarities with the Atomo framework \cite{WSLCPW}, which is a general sparsification method applied on any atomic decomposition. In particular, Kashin representation $x=Ua$ can be viewed as an atomic decomposition of $x\in\R^d$ in lifted space $\R^D$. However, in contrast to Atomo framework, we do not apply sparsification afterwards as it violates the condition (\ref{quantizer_di}). Instead, we utilize quantizers (e.g. coordinate-wise quantization operators) to satisfy (\ref{quantizer_di}), which ensures uniform and dimension-free upper bound for the variance.

\section{Measure concentration and orthogonal matrices}\label{sec:mes-conc}

The concentration of the measure is a remarkable high-dimensional phenomenon which roughly claims that a function defined on a high-dimensional space and having small oscillations takes values highly concentrated around the average \citep{Ledoux-MC,Milman-MC}. Here we present one example of such concentration for Lipschitz functions on the unit sphere, which will be the key to justify the restricted isometry property.

\subsection{Concentration on the sphere for Lipschitz functions}

We say that $f\colon \sphere^{d-1}\to\R$ is a Lipschitz function with constant $L>0$ if for any $x,\,y\in\sphere^{d-1}$
$$
|f(x) - f(y)| \le L\|x - y\|_2.
$$
The following result is a reformulation of Theorem 5.1.4 in \cite{Versh} with explicit absolute constants.

\begin{theorem}\label{thm:conc-lip}
Let $X\in\sphere^{d-1}$ be a random vector uniformly distributed on the unit Euclidean sphere. If $f\colon \sphere^{d-1}\to\R$ is $L-$Lipschitz function, then for any $t\ge0$
$$
\Pr\left(|f(X)-\E f(X)| \ge t\right) \le 5\exp\(-\frac{(d-2)t^2}{8 L^2}\).
$$
\end{theorem}
Informally and rather surprisingly, Lipschitz functions on a high-dimensional unit sphere are almost constants. Particularly, it implies that deviations of function values from the average are at most $\frac{8L}{\sqrt{d}}$ with confidence level more than 0.99. We will apply this concentration inequality for the function $x\to\|U x\|_2$ which is $1-$Lipschitz if $U$ is orthogonal.

\subsection{Random orthogonal matrices}

Up to this point we did not discuss how to choose the frame vectors $u_i$ or the frame matrix $U$, which is used in the construction of Kashin's representation. We only know that it should be orthogonal and satisfy RIP for some parameters $\delta, \eta$. We now describe how to construct frame matrix $U$ and how to estimate parameters $\delta, \eta$. Unluckily, there is no an explicit construction scheme for such matrices. There are random generation processes that provide probabilistic guarantees \citep{Candes-Tao-1,Candes-Tao-2,LyVe}.

Consider random $d\times D$ matrices with orthonormal rows. Such matrices are obtained from selecting the first $d$ rows of orthogonal $D\times D$ matrices. Let $O(D)$ be the space of all orthogonal $D\times D$ matrices with the unique translation invariance and normalized measure, which is called Haar measure for that space. Then the space of $d\times D$ orthogonal matrices is $$O(d\times D) = \{U = P_d V \colon V\in O(D)\},$$ where $P_d\colon\R^D\to\R^d$ is the orthogonal projection on the first $d$ coordinates. The probability measure on $O(d\times D)$ is induces by the Haar measure on $O(D)$. Next we show that, with respect to the normalized Haar measure, randomly generated orthogonal matrices satisfy RIP with high probability. The following result is refinement of Theorem 4.1 in \cite{LyVe}.

\begin{theorem}\label{thm:prob-rip}
Let $\lambda>1$ and $D=\lambda d$, then with probability at least
\begin{equation*}
 1 - 5 \exp\left[-d \left(\sqrt{\lambda} - 1\right)^2 \left(\frac{1}{26} + \frac{1}{208}\log\left(1 - \frac{1}{\sqrt{\lambda}}\right) \right)   \right],
\end{equation*}

a random orthogonal $d\times D$ matrix $U$ satisfies RIP with parameters
\begin{equation}\label{eta-delta-formulas}
 \eta = \frac{3}{4} + \frac{1}{4}\cdot\frac{1}{\sqrt{\lambda}}, \qquad \delta = \frac{1}{5^4}\left(1 - \frac{1}{\sqrt{\lambda}}\right)^2.
\end{equation}
\end{theorem}

Note that the expression for the probability can be negative if $\lambda$ is too close to 1. Specifically, the logarithmic term vanishes for $\lambda \approx 1.0005$ giving negative probability. However, the probability approaches to $1$ quite rapidly for bigger $\lambda$'s. To get a sense of how high that probability can be, note that for $d=1000$ variables and $\lambda=2$ inflation it is bigger than $0.98$.

Now that we have explicit formulas for the parameters $\delta$ and $\eta$, we can combine it with the results of Section \ref{sec:kashin_repr} and summarize with the following theorem.

\begin{theorem}\label{thm:kashin_summary}
Let $\lambda > 1$ be the redundancy factor and $\cC$ be any unbiased compression operator satisfying (\ref{quantizer_di}). Then Kashin Compression $\cC_{\kappa}\in\U(\omega_{\lambda})$ is an unbiased compression with dimension independent variance
\begin{equation}\label{variance-kashin}
 \omega_{\lambda} = \(\frac{10\sqrt{\lambda}}{\sqrt{\lambda}-1}\)^4.
\end{equation}
\end{theorem}

\section{Experiments}

In this section we describe the implementation details of KC and present our experiments of KC compared to other popular compression methods in the literature.

\subsection{Implementation details of KC}

To generate a random (fat) orthogonal frame matrix $U$, we first generate a random matrix with entries drown independently from  Gaussian distribution. Then we extract an orthogonal matrix by applying QR decomposition. Note that, for big dimensions the generation process of frame matrix $U$ becomes computationally expensive. However, after fixing the dimension of to-be-compressed vectors then the frame matrix needs to be generated only once and can be used throughout the learning process.

Afterwards, we turn to the estimation of the parameters $\delta$ and $\eta$ of RIP, which are necessary to compute Kashin's representations.
These parameters are estimated iteratively so to minimize the representation level $K$ (\ref{kashin_level}) subject to the constraint (\ref{up_matrix}) of RIP.
For fixed $\delta$ we first find the least $\eta$ such \ref{up_matrix} holds for unit vectors, which were obtained by normalizing Gaussian random vectors (we chose sample size of $10^4-10^5$, which provided a good estimate). Then we tune the parameter $\delta$ (initially chosen $0.9$) to minimize the level $K$ (\ref{kashin_level}).

\subsection{Empirical variance comparison}

We empirically compare the variance produced by natural dithering \citep{Cnat} against KC with natural dithering and observe that latter introduces much less variance. We generated $n$ vectors with $d$ independent entries from standard Gaussian distribution. Then we fix the minimum number of levels $s$ that allows obtaining an acceptable variance for performing KC with natural dithering. Next, we adjust levels $s$ for natural dithering to the almost same number of bits used for transmission of the compressed vector.
For each of these vectors we compute normalized empirical variance via
\begin{equation}\label{eq:empirical_variance}
 \omega(x)\eqdef \frac{\|\mathcal{C}(x)-x\|^{2}}{\|x\|^{2}}.
\end{equation}

\begin{figure}[t]
  \centering
  \includegraphics[width=0.8\linewidth]{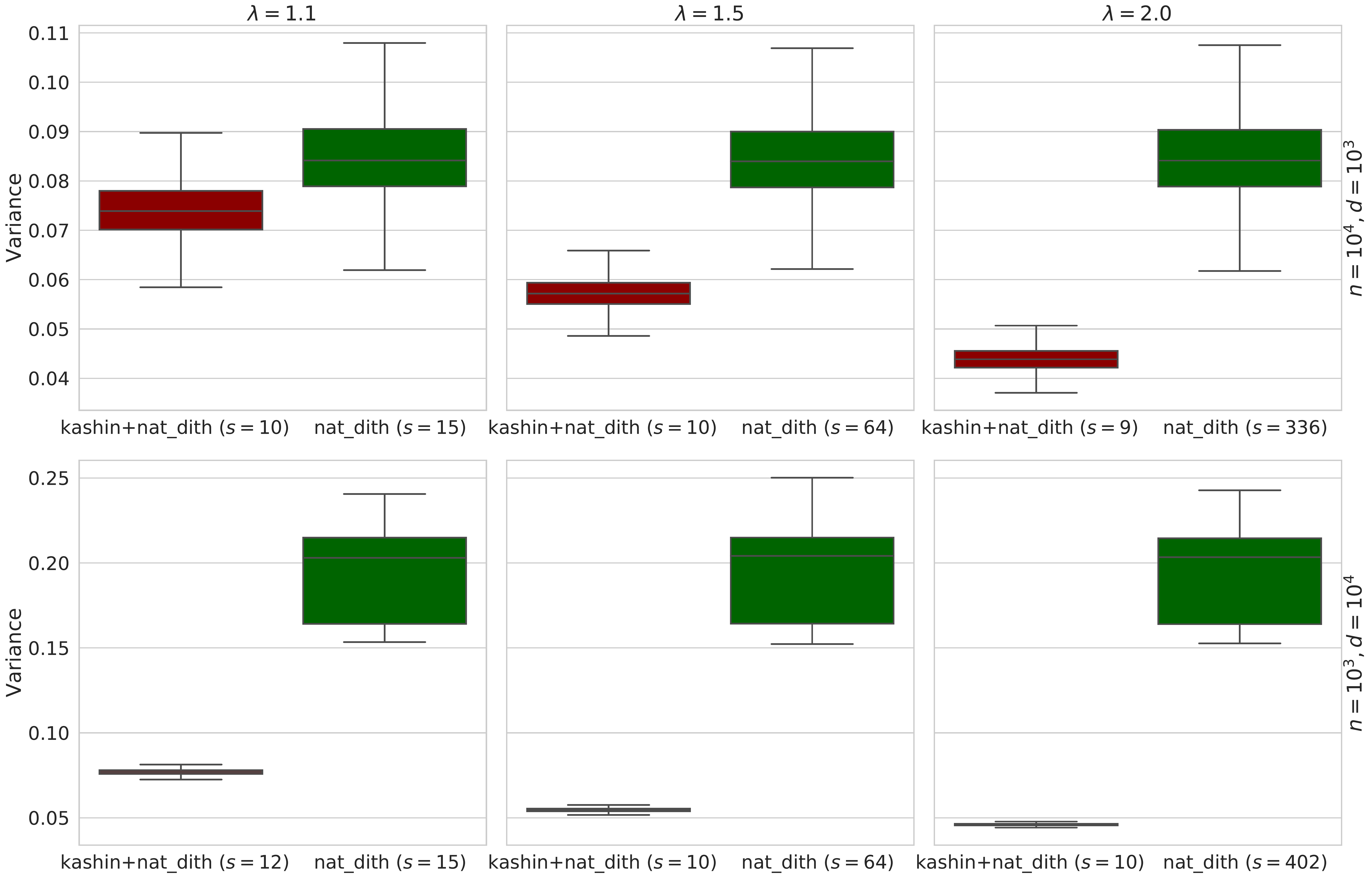}
  \caption{Comparison of empirical variances (\ref{eq:empirical_variance}) of natural dithering and KC with natual dithering.}  
  \label{fig:emp_var_dith_boxplot}
\end{figure}

\begin{figure}[th]
  \centering
  \includegraphics[width=0.8\linewidth]{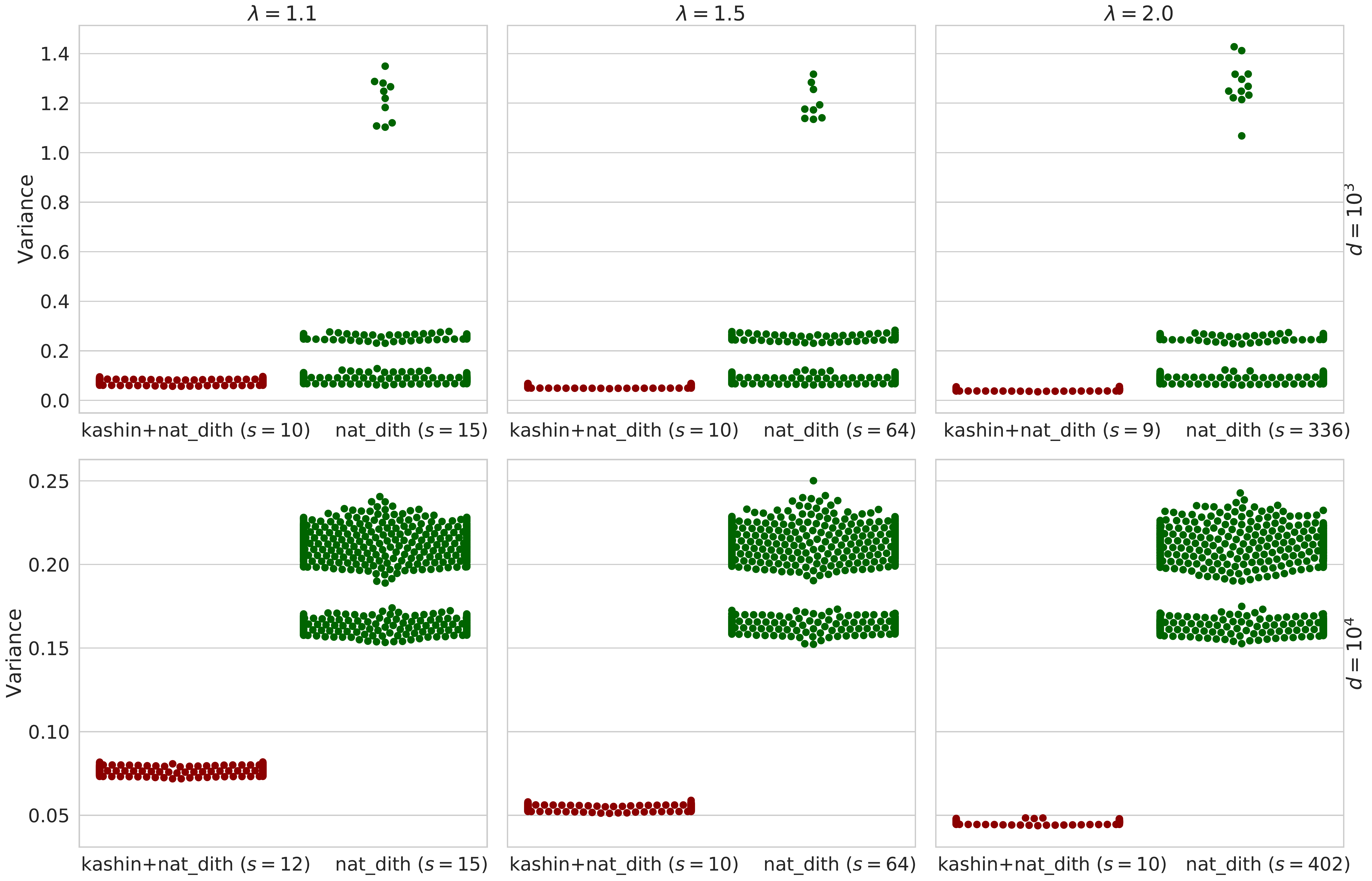}
 \caption{Swarmplots (with sub-sample size $n=1000$) of empirical variances (\ref{eq:empirical_variance}) for natural dithering and KC with natural dithering.}
  \label{fig:emp_var_dith_swarmplot}
\end{figure}

In Figure~\ref{fig:emp_var_dith_boxplot} we provide boxplots for empirical variances, which show that the increase of parameter $\lambda$ leads to smaller variance for KC. They also confirm that for natural dithering, the variance $\omega$ scales with the dimension $d$ while for KC that scaling is significantly reduced (see also Table \ref{table:comp-methods} for variance bounds). This shows the positive effect of KC combined with other compression methods.
For additional insights, we present also swarmplots provided by \href{https://seaborn.pydata.org/generated/seaborn.swarmplot.html}{Seaborn Library}.
Figure~\ref{fig:emp_var_dith_swarmplot} illustrates the strong robustness property of KC with respect to outliers.

\subsection{Minimizing quadratics with CGD}
To illustrate the advantages of KC in optimization algorithms, we minimized randomly generated quadratic functions (\ref{eq:quadratic}) for $d=10^4$ using gradient descent with compressed gradients.
\begin{equation}\label{eq:quadratic}
 \min\limits_{x \in \mathbb{R}^d}~~f(x) = \frac{1}{2} x^{\top} A x - b^{\top} x,
\end{equation}

\begin{figure*}[th]
\centering
\begin{subfigure}{\textwidth}
  \centering
  \includegraphics[width=0.35\linewidth]{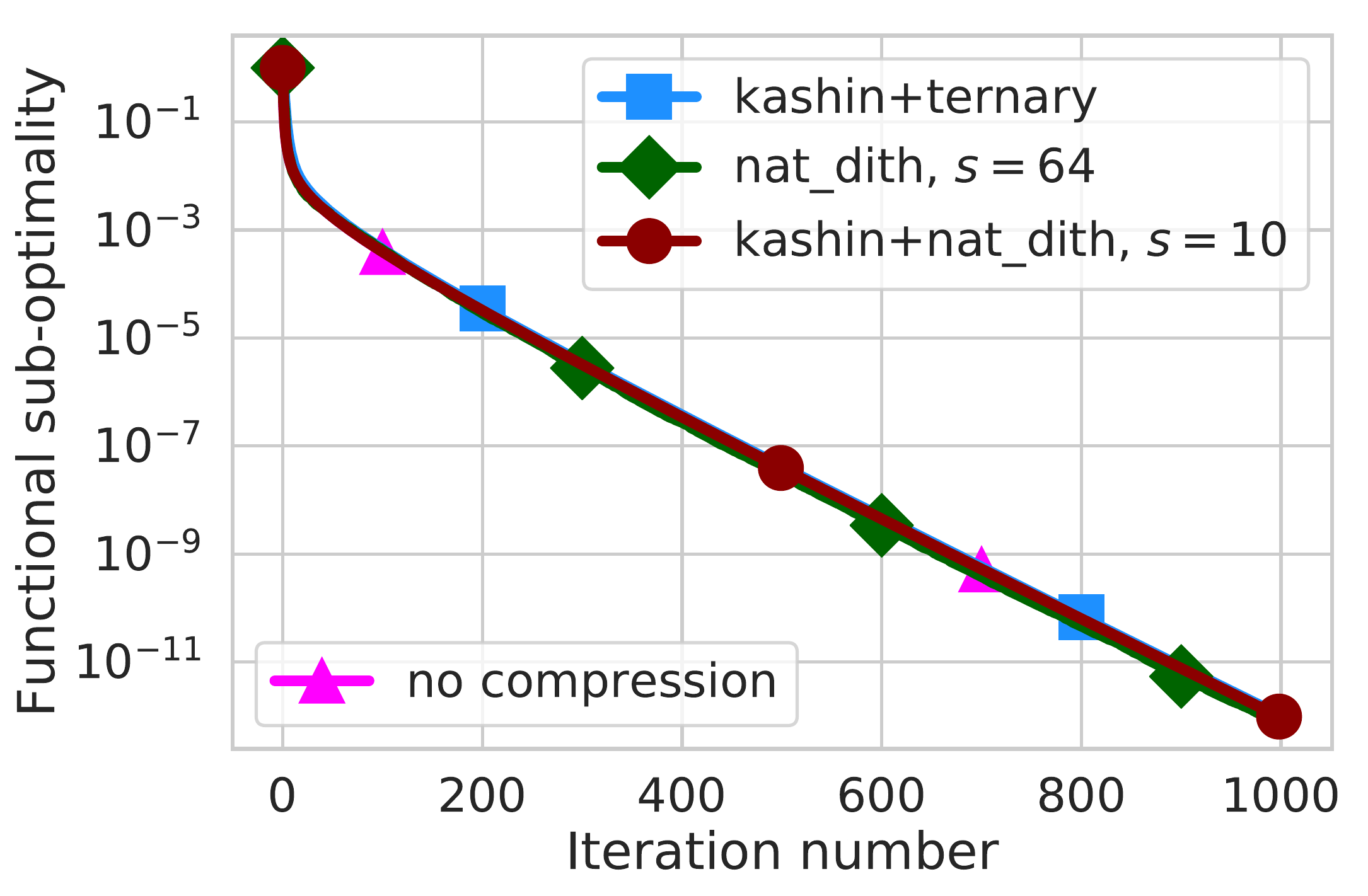}
  \includegraphics[width=0.35\linewidth]{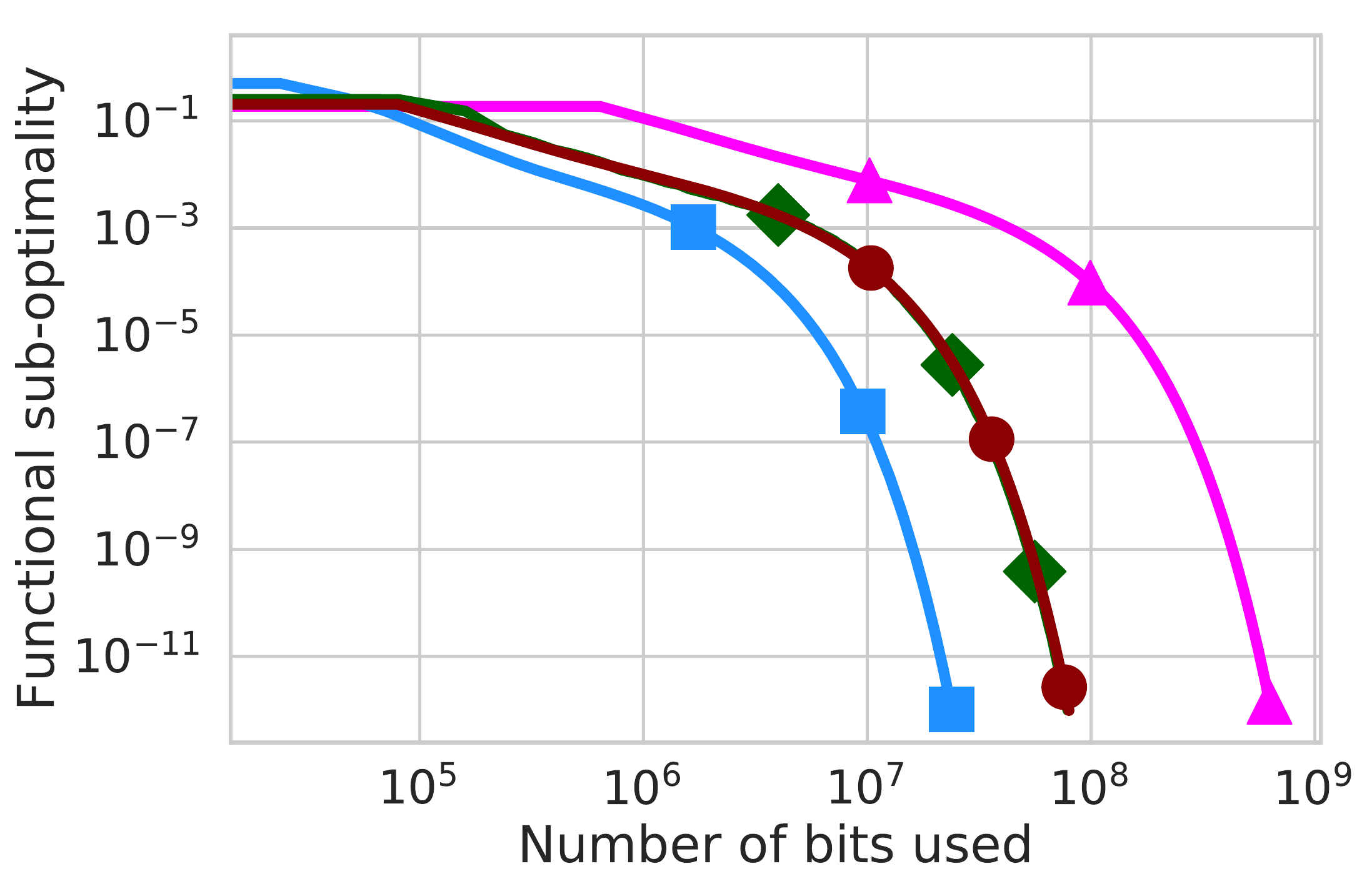}
  \caption{Convergence speeds with respect to the number of gradient steps and amount of communicated bits.} 
  \label{fig:quad_problem}
\end{subfigure}
\begin{subfigure}{\textwidth}
  \centering
  \includegraphics[width=0.35\linewidth]{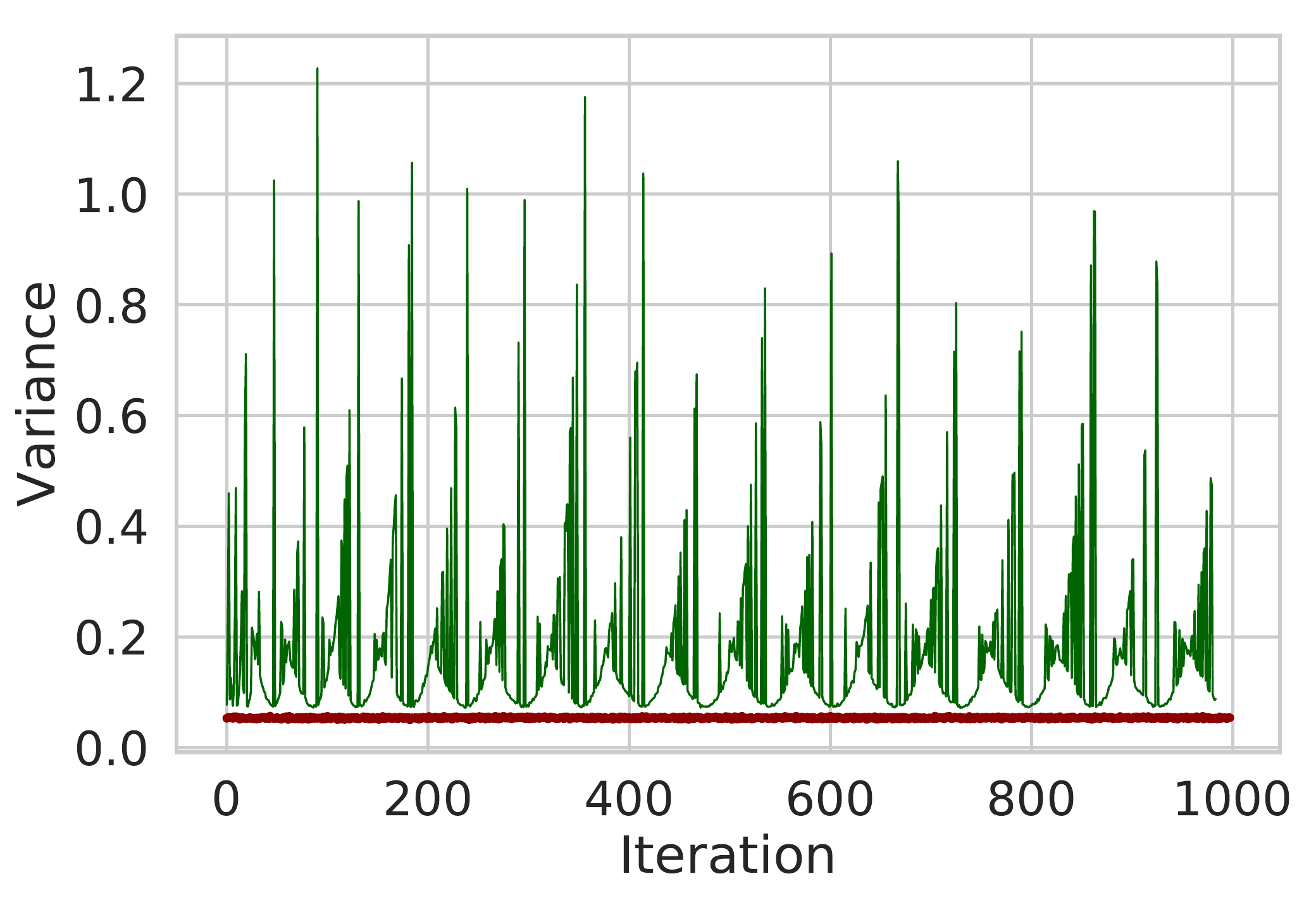}
  \includegraphics[width=0.35\linewidth]{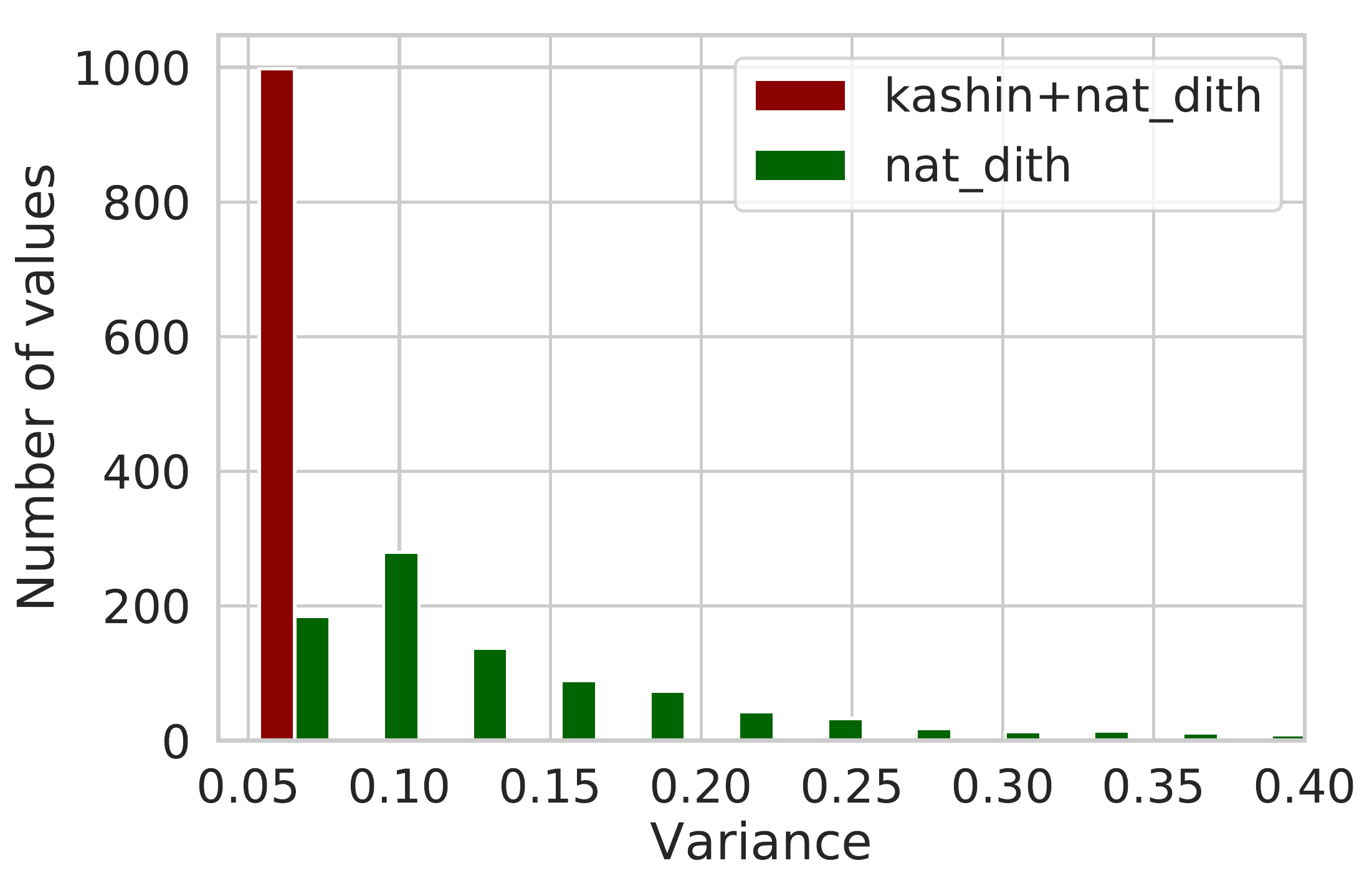}
   \caption{Empirical variances of compressed gradients throughout the optimization process.}
  \label{fig:quad_problem_var}
\end{subfigure}
  \caption{Performance of different compression methods during the minimization of quadratics~(\ref{eq:quadratic}). Hyperparameters of compression operators ($\lambda$ for KC and $s$ for natural dithering) were chosen in such a way so to have either identical function suboptimalities (\ref{fig:quad_problem}) or an identical number of compressed bits (\ref{fig:quad_problem_var}).}
\end{figure*}

In Figure~\ref{fig:quad_problem} we evaluate functional suboptimality $$\frac{f(x_k)-f^*}{f(x_0)-f^*}$$ in log-scale for vertical axis. These plots illustrate
the superiority of KC with ternary quantization, where it does not degrade the convergence at all and saves in communication compared to other compression methods and without any compression scheme.

To provide more insights into this setting, Figure~\ref{fig:quad_problem_var} visualizes empirical variances of the compressed gradients throughout the optimization process, revealing both the low variance feature and the stabilization property of KC. 

\subsection{Minimizing quadratics with distributed CGD}

Consider the minimization problem of the average of $n$ quadratics
\begin{equation}\label{eq:distr_quad}
\min_{x\in\R^d} f(x) \eqdef \frac{1}{n} \sum_{i=1}^n f_i(x), \qquad \text{where} \qquad f_i(x) = \frac{1}{2} x^\top A_i x,
\end{equation}
with synthetically generated matrices $A_i$. We solve this problem with Distributed Compressed Gradient Descent  (Algorithm~\ref{alg:dcgd}) using a selection of compression operators.

 \begin{algorithm}[t]
    \caption{Distributed Compressed Gradient Descent (DCGD)}
    \label{alg:dcgd}
 \begin{algorithmic}
    \STATE {\bfseries Input:} learning rate $\gamma > 0$, starting point $x^{0} \in \mathbb{R}^{d}$, compression operator $\mathcal{C} \in \mathbb{B}(\alpha)$.
    \STATE {\bf for} $k=0,1,2, \ldots$ {\bf do}
    \STATE \quad {\bf for all nodes} $i \in \{1, 2, \ldots, n\}$ {\bf in parallel do}
    \STATE \qquad Compute  local gradient $\nabla f_{i}\(x^{k}\)$ 
    \STATE \qquad Compress local gradient $g_{i}^{k} = \mathcal{C}\left(\nabla f_{i}\left(x^{k}\right)\right)$ \\
    \STATE \qquad Receive the aggregate $g^{k} = \frac{1}{n} \sum \limits_{i=1}^{n} g_{i}^{k}$ \\
    \STATE \qquad $x^{k+1} = x^{k} - \gamma g^{k}$
 \end{algorithmic}
 \end{algorithm}

\begin{figure*}[th!]
  \centering
  \begin{subfigure}{0.35\linewidth}
  \includegraphics[width=\linewidth]{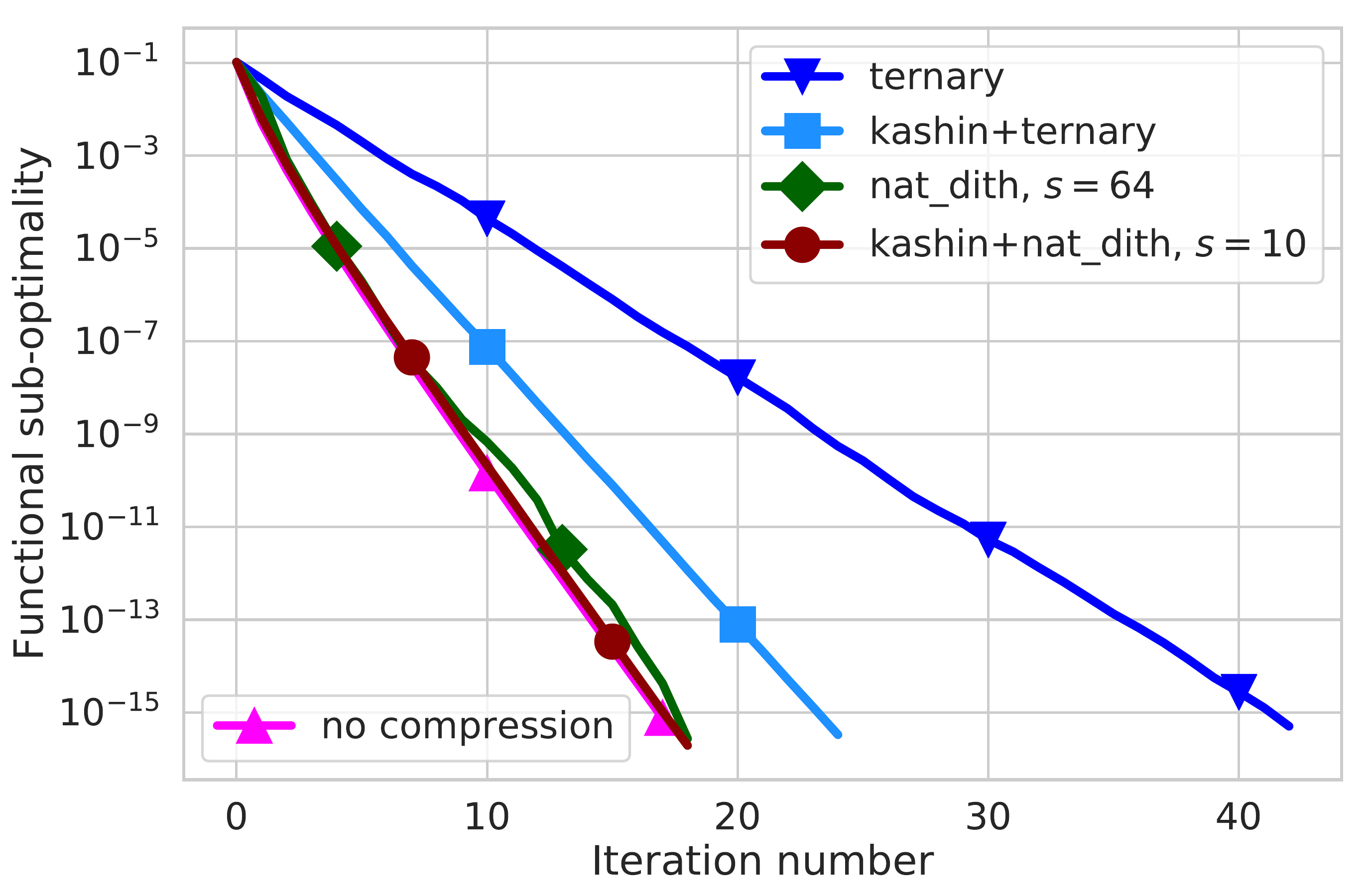}
  \end{subfigure}
  \begin{subfigure}{0.35\linewidth}
    \includegraphics[width=\linewidth]{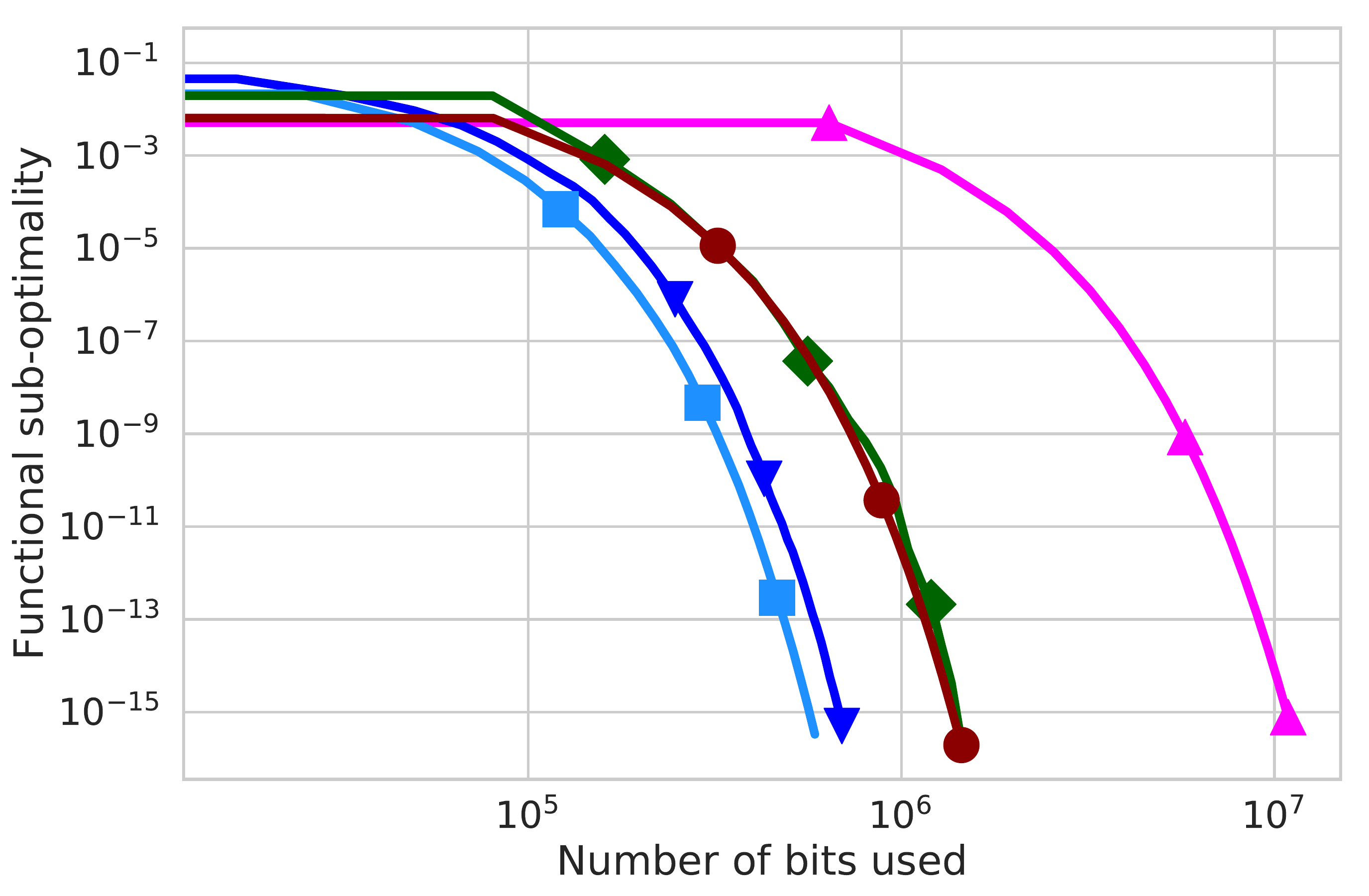}
  \end{subfigure}
  \caption{Performance of Distributed Compressed Gradient Descent (Algorithm~\ref{alg:dcgd} with different compression operators for problem~\eqref{eq:distr_quad} with $n=10$ workers and $d = 10^3$.} 
  \label{fig:distr_10_quad_1000}
\end{figure*}

 \begin{figure*}[th!]
  \centering
  \begin{subfigure}{0.35\linewidth}
  \includegraphics[width=\linewidth]{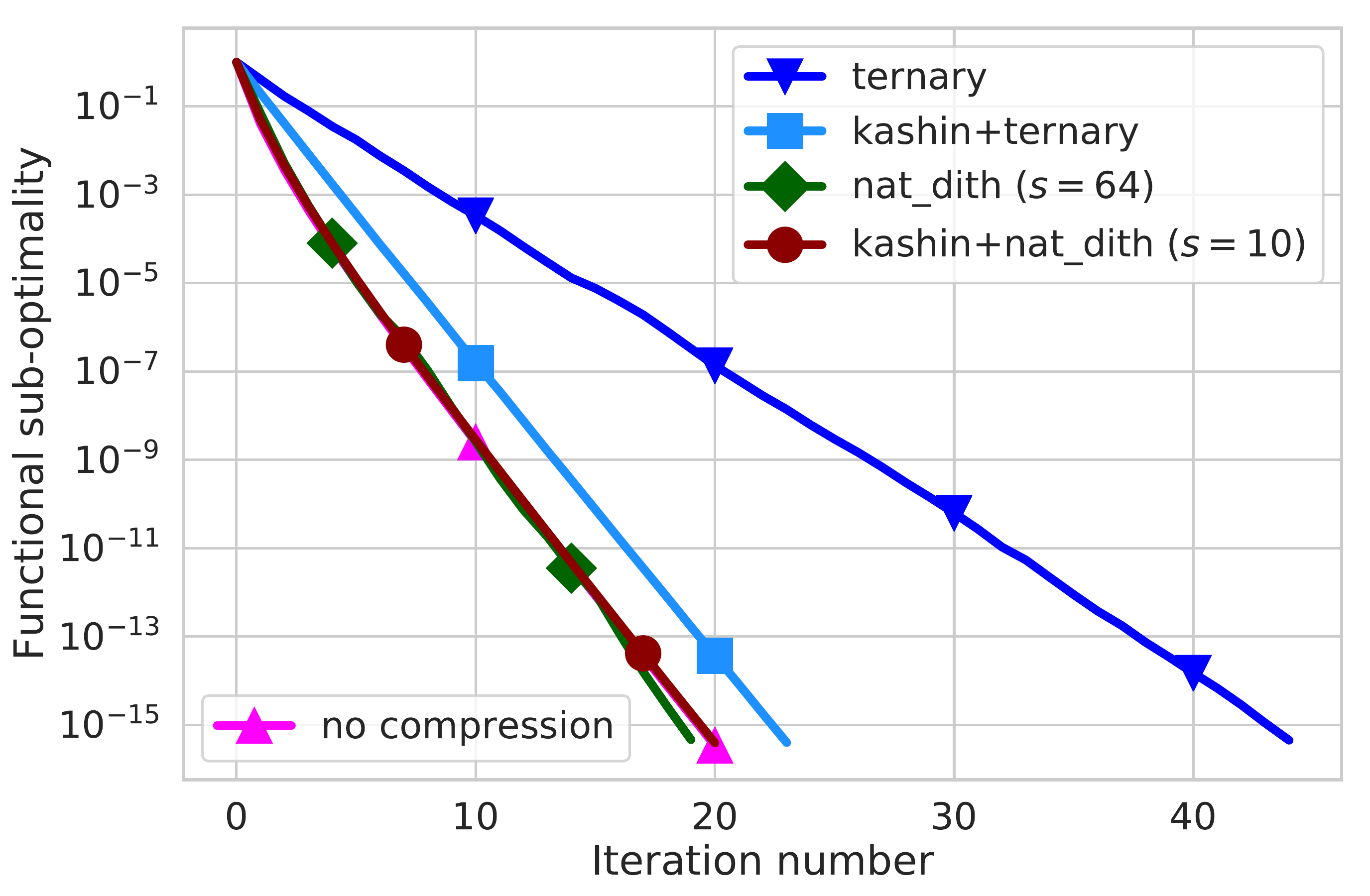}
  \end{subfigure}
  \begin{subfigure}{0.35\linewidth}
    \includegraphics[width=\linewidth]{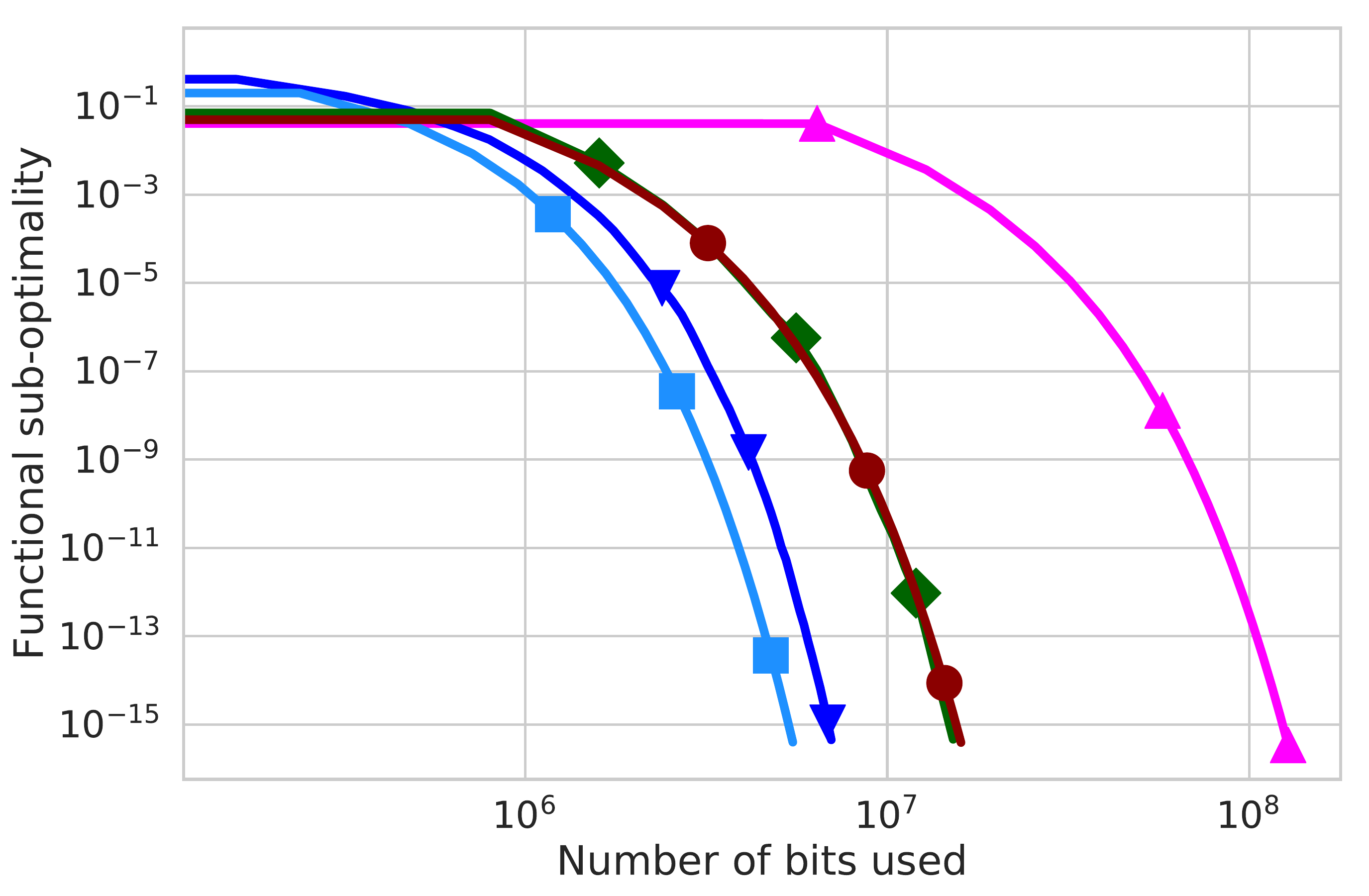}
  \end{subfigure}
  \caption{Performance of Distributed Compressed Gradient Descent (Algorithm~\ref{alg:dcgd}) with different compression operators for problem~\eqref{eq:distr_quad} with $n=10$ workers and $d = 10^4$.} 
  \label{fig:distr_10_quad_10000}
\end{figure*}

Figures~\ref{fig:distr_10_quad_1000} and~\ref{fig:distr_10_quad_10000} show that KC combined with ternary quantization leads to faster convergence and uses less bits to communicate than ternary quantization alone. Note that in higher dimension the gap between KC with ternary quantization and no compression gets smaller in the iteration plot, while in the communication plot it gets bigger. So, in high dimensions KC convergences slightly worse than no compression scheme, but the savings in communication are huge.

\clearpage
\bibliography{references}
\bibliographystyle{dinat}

\clearpage
\appendix
\part*{Appendix}

\section{Proofs for Section \ref{sec:up}}

\subsection{Proof of Theorem \ref{thm:biased_up}: UP for biased compressions $\B(\alpha)$}

Fix $R>0$ and let $B^d(R)$ be the $d$-dimensional Euclidean closed ball with center at the origin and with radius $R$. Denote by $m = 2^b$ the number of possible outcomes of compression operator $\cC$ and by $\{v_1,\dots,v_m\}\subset\R^d$ the set of compressed vectors.
We relax the $\alpha$-contractive requirement and prove (\ref{biased_up}) in the case when the restricted compression operator $\cC\colon B^d(R)\to \{v_1,\dots,v_m\}$ satisfies
\begin{equation}\label{class-biased-relaxed}
\E\[\|\cC(x)-x\|^2\] \le \alpha R^2, \quad x\in B^d(R).
\end{equation}

Define probability functions $p_k$ as follows
$$
p_k(x) = \Pr\(\cC(x) = v_k\), \quad x\in B^d(R),\quad k\in[m].
$$

Then we stack functions $p_k$ together and get a vector valued function $p\colon B^d(R)\to\Delta^m$, where $\Delta^m$ is the standard $m$-simplex
\begin{equation*}
\Delta^m = \left\{(p_1, p_2, \dots, p_m)\in\R^m \colon \sum_{k=1}^m p_k = 1,\, p_k\ge 0\,\text{ for all } k\in[m] \right\}.
\end{equation*}

We can express the expectation in (\ref{class-biased-relaxed}) as
\begin{equation}\label{biased_ineq}
\E\[\|\cC(x)-x\|^2\] = \sum_{k=1}^m p_k(x)\|v_k-x\|^2
\end{equation}
and taking into account the inequality (\ref{class-biased-relaxed}) itself, we conclude
\begin{equation*}
\max_{x\in B^d(R)} \sum_{k=1}^m p_k(x)\|v_k-x\|^2 \le \alpha R^2.
\end{equation*}

The above inequality holds for the particular probability function $p$ defined from the compression $\mathcal{C}$. Therefore the inequality will remain valid if we take the minimum of left hand side over all possible probability functions $\hat{p}\colon B^d(R)\to\Delta^m$:
\begin{equation}\label{bound_01}
\min_{\hat{p}\colon B^d(R)\to\Delta^m} \max_{x\in B^d(R)} \sum_{k=1}^m \hat{p}_k(x)\|v_k-x\|^2 \le \alpha R^2.
\end{equation}

We then swap the order of min-max by adjusting domains properly:
$$
\min_{\hat{p}\colon B^d(R)\to\Delta^m} \max_{x\in B^d(R)} \sum_{k=1}^m \hat{p}_k(x)\|v_k-x\|^2 = \max_{x\in B^d(R)} \min_{\hat{p}\in\Delta^m} \sum_{k=1}^m \hat{p}_k\|v_k-x\|^2,
$$
where the second minimum is over all probability vectors $\hat{p}\in\Delta^m$ (not over vector valued functions as in the first minimum). Next, notice that
$$
\min_{\hat{p}\in\Delta^m} \sum_{k=1}^m \hat{p}_k\|v_k-x\|^2 = \|v_x-x\|^2,
$$
where $v_x \in \arg\min_{v\in\{v_1,\dots,v_m\}} \|v-x\|^2$ is the closest $v_k$ to $x$. Therefore, we have transformed (\ref{bound_01}) into
$$
\max_{x\in B^d(R)} \|v_x-x\| \le R \sqrt{\alpha} =: \hat{R}.
$$

The last inequality means that the set $\{v_1,\dots,v_m\}$ is an $\hat{R}$-net for the ball $B^d(R)$. Using simpler version of the argument on covering numbers and volume (see Proposition 4.2.12, \citep{Versh} for the general case) we conclude
$$
m = \# \{v_1,\dots,v_m\} \ge \frac{\textrm{vol}(B^d(R))}{\textrm{vol}(B^d(\hat{R}))} = \frac{R^d}{\hat{R}^d} = \alpha^{-\nicefrac{d}{2}},
$$
which completes the proof since
$$
\alpha \cdot 4^{\nicefrac{b}{d}} = \alpha \cdot m^{\nicefrac{2}{d}} \ge 1.
$$

\subsection{Proof of Theorem \ref{thm:biased_up}: Derivation from Rate Distortion Theory}\label{apx:up-rd}

After the first online appearance of the work, we had been aware of by an anonymous reviewer that this result follows from the rate distortion theory. We include the derivation here.

Here we deduce the lower bound (\ref{biased_up}) from the results of rate distortion theory, which is a subfield of information theory \citep{Elements_IT}. Rate distortion theory describes theoretical limitations of lossy compression of a given random variable in terms of information rate $R$ and distortion threshold $D$. In our notation, the rate $R=\nicefrac{b}{d}$ is the average number of bits over coordinates and distortion $D = \alpha\sigma^2$, where $\sigma^2$ is the variance of the random variable. Rate distortion function $R(D)$ of a given random source is the minimal rate to transfer an i.i.d. sample such that the receiver can decode the initial sequence with distortion $D$. We assume that squared error distortion measure is used, namely
$$
\rho(x,\hat{x}) = \frac{1}{d}\sum_{i=1}^d (x_i-\hat{x}_i)^2.
$$
This distortion measure is particularly convenient for our setup as inequality (\ref{class-biased}) can be written as
\begin{equation}\label{class-biased-rd}
\E_{\cC}\[\rho(\cC(x),x)\] \le \alpha \rho(x,0), \quad x\in\R^d.
\end{equation}
If (\ref{class-biased-rd}) holds uniformly for any $x\in\R^d$, then it also holds in expectation with respect to $x\sim\cD$ with i.i.d. coordinates sampled from some distribution $\cD$:
$$
\E_{\cC,\cD}\[\rho(\cC(x),x)\] \le \alpha \E_{\cD}\[\rho(x,0)\] = \alpha\sigma^2 = D, \quad x\sim\cD.
$$
This implies that the rate $R$ of compression $\cC$ is bigger than rate distortion function $R(D)$ for any distribution $\cD$. In particular, $R\ge R_{\cN}(D)$ when distribution $\cD$ is Gaussian.
It is known that rate distortion function $R_{\mathcal{N}}(D)$ for Gaussian random variable can be written analytically as
$$
R_{\cN}(D) = 
\begin{cases}
\log_4 \frac{\sigma^2}{D}   & \text{if}\quad 0 \le D \le \sigma^2 \\
0                  & \text{if}\quad D > \sigma^2.
\end{cases}
$$
Translating inequality $R\ge R_{\cN}(D)$ into the language of $\alpha=\nicefrac{D}{\sigma^2}\in[0,1]$ and $b=Rd$, we get $\alpha\cdot 4^{\nicefrac{b}{d}} \ge 1$.

It is worth to mention that Gaussian random variable is the hardest source to encode, meaning it requires the most number of bits to ensure a given distortion constraint. Formally, for any random variable with the same $\sigma^2$ variance the rate distortion function $R(D)\le R_{\cN}(D)$.

\begin{remark}
In practice we do not deal with too large or too small values because of the finite bit representation of a single float in a machine. Therefore, the quantifier $\forall x\in\R^d$ in (\ref{class-biased}) or (\ref{class-biased-rd}) and the idealized Gaussian source used for the lower bound should be relaxed by excluding vectors with too small or too large norms and considering approximate Gaussian distributions.
\end{remark}

\subsection{Proof of Lemma \ref{lem:inclusion}}

Let $\mathcal{C}\in\U(\omega)$. Using relations $\E\[\cC(x)\]=x$ and $\E\[\left\|\cC(x)\right\|^2\] \le (\omega+1)\|x\|^2$, we get
\begin{align*}
\E\[\left\|\frac{1}{\omega+1}\mathcal{C}(x)-x\right\|^2\]
&= \frac{1}{(\omega+1)^2}\E\[\|\mathcal{C}(x)\|^2\] - \frac{2}{\omega+1}\E\[\<\mathcal{C}(x),x\>\] + \|x\|^2 \\
&= \frac{1}{(\omega+1)^2}\E\[\|\mathcal{C}(x)\|^2\] + \(-\frac{2}{\omega+1}+1\)\|x\|^2 \\
&\le \(\frac{1}{\omega+1} -\frac{2}{\omega+1} + 1\)\|x\|^2
= \frac{\omega}{\omega+1}\|x\|^2,
\end{align*}
which concludes the lemma.

\section{Proof for Section \ref{sec:polytop}}

\subsection{Proof of Theorem \ref{asym-tightness}: Asymptotic tightness of UP}

We use the construction of \cite{Koch2}, which states the following:
\begin{theorem}[\cite{Koch2}]\label{kochol-construction}
For any dimension $d\ge 1$ and any fixed $2d\le m \le 2^d$ there exist $m$ unit vectors such that the convex hull of them (or the polytope with such vertices) contains a ball of radius
$$
c_1 \sqrt{\frac{1}{d}\log\frac{m}{d}},
$$
for some absolute constant $c_1>0$.
\end{theorem}

Fix $c\in(0,1)$ and $d\ge d_0$ where $d_0=d_0(c)$ is the smallest possible $d$ such that $2d\le 2^{c d}$.
Choose $m=2^{c d}$ be the number of vertices of the polytope obtained by inversing (with respect to the unit sphere) the $m$ unit vectors from Theorem \ref{kochol-construction}. Clearly, $2d\le m\le 2^d$ as $d\ge d_0$.
Therefore, the obtained polytope contains the unit ball $B^d(0,1)$ and vertices have the same magnitude $R$ satisfying
$$
1 < R \le \frac{1}{c_1} \sqrt{\frac{d}{\log\frac{m}{d}}}.
$$
This construction yields an $\omega$-compressor $\cC\colon\sphere^{d-1}\to\R^d$ with $b=c d$ bits and
$$
\omega + 1 = R^2 \le \frac{1}{c_1^2} \cdot \frac{d}{\log\frac{m}{d}} = \frac{1}{c_1^2} \cdot \frac{d}{cd - \log d}.
$$
Therefore
\begin{equation}\label{temp-01}
\frac{\omega}{\omega+1} \cdot 4^{\nicefrac{b}{d}} = \(1-\frac{1}{\omega+1}\) \cdot 4^{\nicefrac{b}{d}} \le \(1-c_1^2\(c-\frac{\log d}{d}\)\) 4^c.
\end{equation}
Notice that choosing small $c\to0$ and large $d\to\infty$ we can make the right hand side of (\ref{temp-01}) arbitrarily close to 1.

\section{Proofs for Section \ref{sec:mes-conc}}

Proofs presented in this section are adjustments of several standard and classical techniques. We present the proofs here to find out hidden absolute constants.

\subsection{Proof of Theorem \ref{thm:conc-lip}: Concentration on the sphere for Lipschitz functions}

Let $\sphere^{d-1}$ be the unit sphere with the normalized Lebesgue measure $\mu$ and the geodesic metric $\dist(x,y) = \arccos\langle x,y\rangle$ representing the angle between $x$ and $y$. Using this metric, we define the spherical caps as the balls in $\sphere^{d-1}$:
\begin{equation*}
B_a(r) = \{x\in\sphere^{d-1} \colon \dist(x,a)\le r \}, \quad a\in\sphere^{d-1},\,r>0.
\end{equation*}

For a set $A\subset\sphere^{d-1}$ and non-negative number $t\ge0$ denote by $A(t)$ the $t$-neighborhood of $A$ with respect to geodesic metric:
\begin{equation*}
A(t) = \left\{x\in\sphere^{d-1} \colon \dist(x,A)\le t\right\}.
\end{equation*} 

The famous result of P.~Levy on isoperimetric inequality for the sphere states that among all subsets $A \subset\sphere^{d-1}$ of a given measure, the spherical cap has the smallest measure for the neighborhood (see e.g. \citep{Ledoux-MC}).
\begin{theorem}[Levy's isoperimetric inequality]
Let $A \subset\sphere^{d-1}$ be a closed set and let $t\ge 0$. If $B=B_a(r)$ is a spherical cap with $\mu(A) = \mu(B)$, then
$$
\mu\(A(t)\) \ge \mu\(B(t)\) \equiv \mu(B_a(r+t)).
$$
\end{theorem}

We also need the following well known upper bound on the measure of spherical caps

\begin{lemma}
Let $t\ge 0$. If $B\subset\sphere^{d-1}$ is a spherical cap with radius $\pi/2-t$, then
\begin{equation}\label{cap-upper-bound}
\mu(B) \le \sqrt{\frac{\pi}{8}}\exp\left(-\frac{(d-2)t^2}{2}\right).
\end{equation}
\end{lemma}

These two results yield a concentration inequality on the unit sphere around median of the Lipschitz function.

\begin{theorem}\label{thm:conc-lip-median}
Let $f\colon \sphere^{d-1}\to\R$ be a $L$-Lipschitz function (w.r.t. geodesic metric) and let $M=M_f$ be its median, i.e.
$$
\mu\left\{x\colon f(x)\ge M\right\} \ge \frac{1}{2} \quad\text{and}\quad \mu\left\{x\colon f(x)\le M\right\} \ge \frac{1}{2}.
$$
Then, for any $t\ge0$
\begin{equation}\label{isop-ineq-median}
\mu\left\{x\colon |f(x)-M|\ge t\right\} \le \sqrt{\frac{\pi}{2}}\exp\(-\frac{(d-2)t^2}{2 L^2}\).
\end{equation}
\end{theorem}

\begin{remark}
Notice that Lipschitzness w.r.t. geodesic metric is weaker than w.r.t. Euclidean metric. This implies that the obtained concentration holds for $L$-Lipschitz function w.r.t. standard Euclidean distance.
\end{remark}

\subsubsection{Proof of Theorem \ref{thm:conc-lip-median}: Concentration around the median}

Without loss of generality we can assume that $L=1$. Denote
$
A_+ = \{x\colon f(x)\ge M\} \quad\text{and}\quad A_- = \{x\colon f(x)\le M\}
$,
so that $\mu(A_{\pm})\ge1/2 = \mu(B_a(\nicefrac{\pi}{2}))$ for some $a\in\sphere^{d-1}$. Then the isoperimetric inequality (\ref{isop-ineq-median}) and the upper bound (\ref{cap-upper-bound}) imply
\begin{align*}
\mu(A_{\pm}^c(t)) = \mu\{x\colon  \dist(x,A_{\pm}) > t \}
&\le \mu\{x\colon  \dist(x,B_a(\nicefrac{\pi}{2})) > t \} \\
&= \mu(B_a(\nicefrac{\pi}{2}-t))
\le \sqrt{\frac{\pi}{8}}\exp\left(-\frac{(d-2)t^2}{2}\right).
\end{align*}
Note that $x\in A_{-}(t)$ implies that $\dist(x,y)\le t,\, f(y)\le M$ for some $y\in A_{-}$. Using the Lipschitzness of $f$ we get $f(x)\le f(y) + \dist(x,y) \le M+t$. Analogously, $x\in A_{+}(t)$ implies that $\dist(x,y)\le t,\, f(y)\ge M$ for some $y\in A_{+}$. Again, the Lipschitzness of $f$ gives $-f(x)\le -f(y) + \dist(x,y) \le -M+t$. Thus
\begin{equation*}
|f(x)-M| \le t \quad\text{for any}\quad x\in A_{+}(t)\cap A_{-}(t).
\end{equation*}
To complete the proof, it remains to combine this with inequalities for measures of complements
\begin{align*}
\mu\(\left\{x\colon |f(x)-M| > t\right\}\)
&= 1 - \mu\(\left\{x\colon |f(x)-M|\le t\right\}\)
\le 1 - \mu(A_{+}(t)\cap A_{-}(t)) \\
&\le \mu(A_{+}^c(t)) + \mu(A_{-}^c(t)) \le \sqrt{\frac{\pi}{2}}\exp\left(-\frac{(d-2)t^2}{2}\right).
\end{align*}
Continuity of $\mu$ and $f$ give the result with the relaxed inequality.


\subsubsection{Proof of Theorem \ref{thm:conc-lip}: Concentration around the mean}

Now, from (\ref{isop-ineq-median}) we derive a concentration inequality around the mean rather than median, where mean is defined via
$$
\E f = \int_{\sphere^{d-1}} f(x)\,d\mu(x).
$$

Again, without loss of generality we assume that $L=1$ and $d\ge 3$. Fix $\epsilon\in[0,1]$ and decompose the set $\left\{x\colon |f(x)-\E f|\ge t\right\}$ into two parts:
$$
\mu\(\left\{x\colon |f(x)-\E f|\ge t\right\}\) \le \mu\(\left\{x\colon |f(x)-M|\ge \epsilon t\right\}\) + \mu\(\left\{x\colon |\E f - M|\ge (1-\epsilon)t\right\}\) =: A_1 + A_2,
$$
where $M$ is a median of $f$. From the concentration (\ref{isop-ineq-median}) around the median, we get an estimate for $A_1$
$$
A_1 \le \sqrt{\frac{\pi}{2}}\exp\(-\frac{(d-2)t^2\epsilon^2}{2}\).
$$
Now we want to estimate the second term $A_2$ with a similar upper bound so to combine them. Obviously, the condition in $A_2$ does not depend on $x$, and it is a piecewise constant function of $t$. Therefore
\begin{align*}
A_2 &\le \mu\(\left\{x\colon \E|f - M|\ge (1-\epsilon)t\right\}\) = \mu\(\left\{x\colon \|f - M\|_1\ge (1-\epsilon)t\right\}\) \\
&=\begin{cases}
1   & \text{if}\quad t \le \nicefrac{1}{(1-\epsilon)}\|f - M\|_1 \\
0   & \text{otherwise}
\end{cases}
\le\begin{cases}
1   & \text{if}\quad t \le \frac{\pi}{2(1-\epsilon)\sqrt{d-2}} \\
0   & \text{otherwise}
\end{cases}
\end{align*}
where we bounded $\|f-M\|_1$ as follows
\begin{align*}
\|f-M\|_1
&= \int_0^{\infty} \mu\(\{x\colon |f(x)-M|\ge u\}\)\,du
\le \sqrt{\frac{\pi}{2}} \int_0^{\infty} \exp\(-\frac{(d-2)u^2}{2}\)\,du \\
&= \sqrt{\frac{\pi}{d-2}} \int_0^{\infty} \exp(-u^2)\,du
= \sqrt{\frac{\pi}{d-2}}\frac{\sqrt{\pi}}{2} = \frac{\pi}{2\sqrt{d-2}}.
\end{align*}

We further upper bound $A_2$ to get the same exponential term as for $A_1$:
\begin{equation}\label{bound-A2}
A_2
\le\begin{cases}
1   & \text{if}\quad t \le \frac{\pi}{2(1-\epsilon)\sqrt{d-2}} \\
0   & \text{otherwise}
\end{cases}
\le
\exp\[\frac{\pi^2}{8}\frac{\epsilon^2}{(1-\epsilon)^2}\] \exp\(-\frac{(d-2)t^2\epsilon^2}{2}\).
\end{equation}

To check the validity of the latter upper bound, first notice that for $t=\frac{\pi}{2(1-\epsilon)\sqrt{d-2}}$ both are equal to 1. Then, the monotonicity and positiveness of the exponential function imply (\ref{bound-A2}) for $0\le t < \frac{\pi}{2(1-\epsilon)\sqrt{d-2}}$ and $t > \frac{\pi}{2(1-\epsilon)\sqrt{d-2}}$.
Combining these two upper bounds for $A_1$ and $A_2$, we get
\begin{equation*}
A_1 + A_2 \le \(\exp\[\frac{\pi^2}{8}\frac{\epsilon^2}{(1-\epsilon)^2}\] + \sqrt{\frac{\pi}{2}}\) \exp\(-\frac{(d-2)t^2\epsilon^2}{2}\) \le 5\exp\(-\frac{(d-2)t^2}{8}\)
\end{equation*}
if we set $\epsilon = \nicefrac{1}{2}$. To conclude the theorem, note that normalized uniform measure $\mu$ on the unit sphere can be seen as a probability measure on $\sphere^{d-1}$.

\subsection{Proof of Theorem \ref{thm:prob-rip}: Random orthogonal matrices with RIP}

Most of the proof follows the steps of the proof of Theorem 4.1 of \cite{LyVe}. First, we relax the inequality in Theorem \ref{thm:conc-lip} to
\begin{equation}\label{ineq-mc-relaxed}
\Pr\(|f(X)-\E f(X)| \ge t\) \le 5\exp\(-\frac{d\, t^2}{9 L^2}\), \quad t \ge 0,\, d\ge 20.
\end{equation}

Let $x\in\sphere^{D-1}$ be fixed. Any orthogonal $d\times D$ matrix $U\in O(d\times D)$ can be represented as the projection $U=P_d V$ of $D\times D$ orthogonal matrix $V\in O(D)$. The uniform probability measure (or Haar measure) on $O(D)$ ensures that if $V\in O(D)$ is random then the vector $z=Vx$ is uniformly distributed on $\sphere^{D-1}$. Therefore, if $U\in O(d\times D)$ is random with respect to the induced Haar measure on $O(d\times D)$, then random vectors $U x$ and $P_d z$ have identical distributions. Denote $f(z)=\|P_d z\|_2$ and notice that $f$ is $1$-Lipschitz on the sphere $\sphere^{D-1}$. To apply the concentration inequality (\ref{ineq-mc-relaxed}), we compute the expected norm of these random vectors:
\begin{eqnarray*}
\E f(z)
&\le & \( \int_{\sphere^{D-1}} \|P_d z\|_2^2\,d\mu(z) \)^{\nicefrac{1}{2}}  =  \( \sum_{i=1}^d \int_{\sphere^{D-1}} z_i^2\,d\mu(z) \)^{\nicefrac{1}{2}} =  \( \sum_{i=1}^d \frac{1}{D} \)^{\nicefrac{1}{2}} = \sqrt{\frac{d}{D}},
\end{eqnarray*}
where we used the fact that coordinates $z_i^2$ are distributed identically and therefore they have the same $\nicefrac{1}{D}$ mean. Applying inequality (\ref{ineq-mc-relaxed}) yields, for any $t\ge 0$
\begin{eqnarray}
\Pr\(U\in O(d\times D) \colon \|U x\|_2 > \sqrt{\nicefrac{d}{D}} + t \) & \le & \Pr\(z\in\sphere^{D-1} \colon |f(z) - \E f(z)| > t \) \notag \\
&\le & 5\exp\(-\frac{D t^2}{9}\).\label{conc-U}
\end{eqnarray}

Let $S^{\delta}$ be the set of vectors $x\in\sphere^{D-1}$ with at most $\delta D$ non-zero elements
\begin{equation*}
S^{\delta} := \left\{x\in\sphere^{D-1} \colon |\supp(x)| \le \delta D\right\} = \bigcup_{|I|\le\delta D} \left\{x\in\sphere^{D-1} \colon \supp(x) \subseteq I\right\} = \bigcup_{|I|\le\delta D} S^{\delta}_I,
\end{equation*}
where $S^{\delta}_I$ denotes the subset of vectors $S^{\delta}$ having a given support $I\subseteq[D]$ of indices.
Fix $\varepsilon>0$. For each $I$, we can find an $\varepsilon$-net for $S^{\delta}_I$ in the Euclidean norm with cardinality at most $\(\nicefrac{3}{\varepsilon}\)^{\delta D}$ (see Proposition 4.2.12 and Corollary 4.2.13 in \citep{Versh}). Taking the union over all sets $I$ with $|I| = \ceil*{\delta D}$, we conclude by the Stirling's approximation that there exists an $\varepsilon$-net $\mathcal{N}_{\varepsilon}$ of $S^{\delta}$ with cardinality
\begin{equation}\label{net-upper-bound}
|\mathcal{N}_{\varepsilon}| \le \binom{D}{\ceil*{\delta D}} \(\frac{3}{\varepsilon}\)^{\delta D} \le \(\frac{3 e}{\varepsilon\delta}\)^{\delta D}.
\end{equation}

Applying inequality (\ref{conc-U}), we have
\begin{equation}\label{prob-bound-over-net}
\Pr\(U\in O(d\times D) \colon \|U y\|_2 > \sqrt{\nicefrac{d}{D}} + t,\,\text{ for some }\,y\in\mathcal{N}_{\varepsilon}\) \le |\mathcal{N}_{\varepsilon}| \cdot 5\exp\(-\frac{D t^2}{9}\).
\end{equation}

Since $\mathcal{N}_{\varepsilon}$ is an $\varepsilon$-net for $S^{\delta}$, then for any $x\in S^{\delta}$ there exists such $y\in\mathcal{N}_{\varepsilon}$ that $\|x-y\|_2 \le \varepsilon$. Furthermore, from the orthogonality of matrix $U$ we conclude
$$
\|U x\|_2 \le \|U y\|_2 + \|U(x-y)\|_2 \le \|U y\|_2 + \varepsilon.
$$

Hence, by relaxing the condition of probability in (\ref{prob-bound-over-net}) and using the upper bound (\ref{net-upper-bound}), we get
\begin{align*}
\Pr\(U\in O(d\times D) \colon \|U x\|_2 > \sqrt{\nicefrac{d}{D}} + t + \varepsilon,\,\text{ for some }\,x\in S^{\delta}\) 
&\le \(\frac{3 e}{\varepsilon\delta}\)^{\delta D} \cdot 5\exp\(-\frac{D t^2}{9}\) \\
&=   5\exp\[-D\(\frac{t^2}{9} - \delta\log{\frac{3 e}{\varepsilon\delta}}\)\].
\end{align*}

The above inequality can be reformulated in terms of RIP condition for a random matrix $U\in O(d\times D)$
\begin{equation}\label{ineq-prob-rip}
\Pr\left(U \in {\rm RIP}\left(\delta, \frac{1}{\sqrt{\lambda}} + t + \varepsilon \right) \right) \ge 1 - 5\exp\[-D\(\frac{t^2}{9} - \delta\log{\frac{3 e}{\varepsilon\delta}}\) \].
\end{equation}

Thus, recalling the formula (\ref{kashin_level}) for the level $K$, we aim to choose such $\varepsilon, t, \delta$ (depending on $\lambda$) that to maximize both $\nicefrac{1}{K}$ and the probability in (\ref{ineq-prob-rip}), i.e. the following two expressions
\begin{equation}\label{terms-to-maximize}
\sqrt{\delta} \left(1 - \frac{1}{\sqrt{\lambda}} - \varepsilon - t\right) \quad\text{and}\quad
\frac{t^2}{9} - \delta\log{\frac{3 e}{\varepsilon\delta}}.
\end{equation}

Note that choosing parameters $\varepsilon, t, \delta$ is not trivial in this case as we want to maximize both terms and there is a trade-off between them. We choose the parameters as follows (these expressions were constructed using two techniques: solving optimality conditions for the Lagrangian and numerical simulations.)
\begin{equation}\label{par_choice}
\varepsilon = \frac{1}{100}\left(1 - \frac{1}{\sqrt{\lambda}}\right), \quad
t = 74\,\varepsilon, \quad
\delta = 16\varepsilon^2.
\end{equation}

With these choice of parameters we establish (\ref{eta-delta-formulas}).
\begin{equation}\label{formulas-delta-eta}
\eta = \frac{1}{\sqrt{\lambda}} + t + \varepsilon = 1 - 25\,\varepsilon = \frac{3}{4} + \frac{1}{4}\cdot\frac{1}{\sqrt{\lambda}}, \quad
\delta = 16 \varepsilon^2 = \frac{1}{5^4}\(1-\frac{1}{\sqrt{\lambda}}\)^2.
\end{equation}

To complete the theorem we need to bound the second expression of (\ref{terms-to-maximize}) for the probability. Letting $\nu = \left(1 - \frac{1}{\sqrt{\lambda}}\right)^2 \in (0,1)$ and plugging the expressions (\ref{par_choice}) in (\ref{terms-to-maximize}) we get
\begin{align*}
\frac{t^2}{9} - \delta\log{\frac{3 e}{\varepsilon\delta}}
&= \frac{74^2}{9}\varepsilon^2 - 16\varepsilon^2\log\frac{\nicefrac{3e}{16}}{\varepsilon^3} \\
&= \frac{74^2}{9\cdot 10^4}\nu - \frac{16}{10^4}\frac{3}{2} \nu \log\frac{(\nicefrac{3e}{16})^{2/3} \cdot 10^4}{\nu} \\
&= A\nu - B\nu\log\frac{C}{\nu}
= (A-B\log C)\nu + B\nu\log\nu
= \nu\left( A-B\log C + B\log\nu \right) \\
&= \left(1 - \frac{1}{\sqrt{\lambda}}\right)^2 \left((A-B\log C) + 2B \log\left(1 - \frac{1}{\sqrt{\lambda}}\right) \right)\\
&\ge \left(1 - \frac{1}{\sqrt{\lambda}}\right)^2 \left(\frac{1}{26} + \frac{1}{208}\log\left(1 - \frac{1}{\sqrt{\lambda}}\right) \right),
\end{align*}
where we defined absolute constants $A, B, C$ as
$$
A = \frac{74^2}{9\cdot 10^4}, \quad B = \frac{24}{10^4}, \quad C = \(\frac{3e}{16}\)^{2/3} \cdot 10^4.
$$
and used the following estimates
$
A-B\log C \ge \frac{1}{26}, 2B = \frac{3}{5^4} \le \frac{1}{208}
$.
This concludes the theorem as
\begin{align*}
\Pr\(U \in {\rm RIP}\(\delta, \eta \)\)
&\ge 1 - 5\exp\[-D\(\frac{t^2}{9} - \delta\log{\frac{3 e}{\varepsilon\delta}}\) \] \\
&\ge 1 - 5\exp\[-D \left(1 - \frac{1}{\sqrt{\lambda}}\right)^2 \left(\frac{1}{26} + \frac{1}{208}\log\left(1 - \frac{1}{\sqrt{\lambda}}\right) \right) \] \\
&\ge 1 - 5\exp\[- d \left(\sqrt{\lambda} - 1\right)^2 \left(\frac{1}{26} + \frac{1}{208}\log\left(1 - \frac{1}{\sqrt{\lambda}}\right) \right) \].
\end{align*}

\subsection{Proof of Theorem \ref{thm:kashin_summary}: Kashin Compression}

The unbiasedness of $\cC_{\kappa}$ has been shown in part \ref{subsec:quant-kashin} with uniform upper bound $K(\lambda)^2$ for the variance. To prove the formula (\ref{variance-kashin}) we use expressions (\ref{formulas-delta-eta})
\begin{equation*}
\omega_{\lambda} = K(\lambda)^2 = \(\frac{1}{\sqrt{\delta}(1-\eta)}\)^2 = \(\frac{1}{4\varepsilon\cdot 25\varepsilon}\)^2 = \(\frac{1}{10\varepsilon}\)^4 = \(\frac{10\sqrt{\lambda}}{\sqrt{\lambda}-1}\)^4.
\end{equation*}

\end{document}